\documentclass{article} 
\usepackage{iclr2027_conference,times}

\usepackage{amsmath,amsfonts,bm}

\def\eqref#1{equation~\ref{#1}}

\def\1{\bm{1}}

\DeclareMathAlphabet{\mathsfit}{\encodingdefault}{\sfdefault}{m}{sl}
\SetMathAlphabet{\mathsfit}{bold}{\encodingdefault}{\sfdefault}{bx}{n}

\usepackage{graphicx}
\usepackage{algorithm}
\usepackage{algpseudocode}
\usepackage{amssymb}
\usepackage{booktabs}
\usepackage{multirow}
\usepackage{subcaption}
\usepackage{hyperref}
\usepackage{url}
\usepackage{float}
\title{Foresight: Planning Future Perception in Streaming VLMs without Retraining
}

\author{
Ashok Prasad Neupane$^{1,*,\dagger}$ \quad
Dipan Bartaula$^{2,*}$ \quad
Ankit Belbase$^{1}$ \quad
Saugat Adhikari$^{1}$ \\
\textbf{Samip Ghimire$^{1}$ \quad
Saroj Poudel$^{1}$ \quad
Binod Bhattarai$^{2,3,4}$ \quad
Danda Pani Paudel$^{2,5}$} \\[0.6em]
\normalfont\small
$^{1}$Independent Researcher \quad
$^{2}$NAAMII, Nepal \quad
$^{3}$University College London, UK \quad
$^{4}$University of Aberdeen, UK \\
$^{5}$INSAIT, Sofia University ``St.~Kliment Ohridski'', Bulgaria \\[0.3em]
$^{*}$Equal contribution \quad
$^{\dagger}$Corresponding author: \texttt{neupane.ashok.9696@gmail.com}
}

\iclrfinalcopy 
\begin{document}

\maketitle
\lhead{Preprint}


\vspace{-0.8em}

\begin{center}
    \captionsetup{type=figure}
    \includegraphics[width=\textwidth]{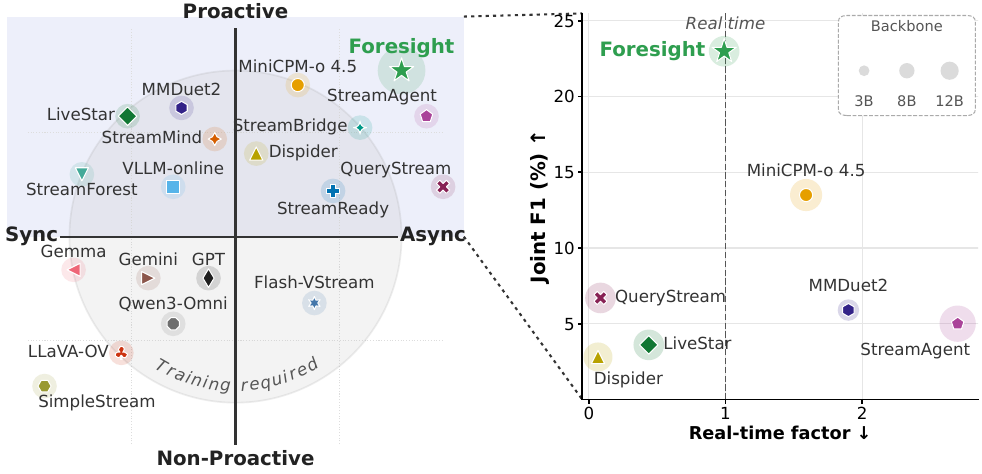}
    \caption{\textbf{Foresight shifts streaming VLMs from reactive to proactive inference}. Left: existing methods often require additional training, where the proactive and asynchronous nature are vital for human-AI interactivity. Right: Foresight performs better than training based methods while being proactive, asynchronous, and superior to real time inference on the challenging OmniPro benchmark.}
    \label{fig:teaser}
    \label{fig:teaser-bubble}
\end{center}

\vspace{0.4em}


\begin{abstract}



Existing streaming vision-language models (VLMs) continuously perceive and
reason over visual streams, but their computational pathways remain fixed
throughout inference. Consequently, they cannot adapt computation to evolving
scene dynamics, where different future events demand different levels and
forms of perception. We show that streaming VLMs inherently possess the ability
to anticipate the immediate future, and leverage this capability to dynamically
configure future computation in a training-free manner. Realizing such
anticipatory computation, however, is very challenging: future anticipation
must be sufficiently reliable to guide computation, planning must run
concurrently with streaming inference, and online reconfiguration must incur
negligible overhead.

To address these challenges, we introduce \textsc{Foresight}, a dual-stream
architecture comprising two Siamese LLMs with shared weights, input encoders,
and KV cache. The first LLM continuously processes incoming tokens, while the
second runs ahead of the stream to anticipate future context, plan future
computation, and generate task responses without interrupting streaming
inference. Each plan decides when to reason next, what to check then, and how
densely to sample, keeping transient evidence separate from persistent control.
The resulting computation plan is executed online through an efficient
reconfiguration protocol with schema-guided decoding and lightweight diff-based
updates, enabling dynamic adaptation with low overhead. With a frozen Qwen3-VL-8B backbone, \textsc{Foresight} achieves 23.0 mean joint F1 on OmniPro Online evaluation beating strongest trained baseline by 9.5\%, while improving the backbone by 6.7 on StreamingBench and 15.4 on OVO-Bench, with the largest gain of 18.7 when evidence arrives later in the video stream. Our source code will be made publicly available.

\end{abstract}

\section{Introduction}
\label{sec:introduction}
Vision-language models (VLMs)~\citep{lin2024vila,li2024llavaonevision,wang2024qwen2vl} have made substantial progress in understanding complex visual content, but most existing models operate offline: they receive a fixed input, perform a predetermined computation, and produce an answer. In contrast, real-world visual intelligence is inherently streaming. An agent observing a live video must continuously process new observations while deciding \emph{when} the available evidence is sufficient, \emph{what} information should be examined next, and \emph{how much} computation should be allocated to the evolving scene. These decisions are especially important in human-AI interaction, where the system must respond to events \emph{proactively}, without being explicitly prompted, and \emph{asynchronously}, staying engaged with the user and the stream while deeper computation runs in the background.

Existing streaming VLMs meet these requirements only in part. Many learn when
to respond
\citep{chen2024videollmonline,wang2025videollm,wang2026mmduet2,qian2025dispider,azad2026streamready},
reduce the cost of each step through frame sampling or token pruning
\citep{chen2024fastv,yao2025timechatonline}, or use memory compression
\citep{qian2024streaminglongvideounderstanding,xu2026streamingvlmrealtimeunderstandinginfinite,chen2026streamtttreconcilingrealtimeperception};
some also separate lightweight stream monitoring from heavier asynchronous
reasoning
\citep{qian2025dispider,wang2025streambridge,ding2025streammind,li2025lionfs,yakushev2025asynchronous}.
Figure~\ref{fig:teaser} sorts these methods according to whether they are
proactive and whether they are asynchronous. Yet even methods that are both
proactive and asynchronous still control computation reactively, based on what
has already arrived rather than on what is expected. In an evolving scene, the
best next step is often to wait for the right evidence. Existing systems leave
that wait to either a fixed schedule \citep{yang2025streamagent} or a learned
trigger \citep{wang2025streambridge,ding2025streammind,azad2026streamready}.

We argue that this wait should be planned by the model itself. In fact, a streaming VLM already understands the scene well enough to know what it is still missing, so it can also estimate when that information will appear and what to check once it does. We therefore treat anticipation not as a prediction target but as a way to control computation: at each reasoning step, the model identifies what remains uncertain, decides when to look again, and sets how densely to sample until then. Streaming inference then becomes a closed loop rather than a fixed sequence of computations, with the model's current state deciding how it processes the future. Figure~\ref{fig:foresight-overview} sketches this loop: one copy of a frozen VLM keeps ingesting the stream, while a second copy reads the same state in the background and writes the plan. Because planning never pauses perception and requires no retraining, this places our approach in the proactive, asynchronous, training-free corner of Figure~\ref{fig:teaser}.

Turning anticipation into control raises three planning challenges: (i) reliability: wrong anticipation misguides reasoning; (ii) concurrency: planning must run alongside perception; and (iii) efficiency: its cost must not negate adaptive inference gains. We address these with \textbf{\textsc{Foresight}}, a training-free perception-planning architecture built on a frozen VLM. For reliability, each plan separates a \emph{transient} evidence state, rebuilt from current observations at every step, from a \emph{persistent} control update that records when to check next and which question to revisit. The responses are further gated by hysteresis on the model's own evidence confidence, so small fluctuations do not trigger premature or repeated outputs. For concurrency, we use a Siamese architecture comprising Ingest and Think copies. The Ingest copy is the only writer to the shared KV cache, while the Think copy plans from a read-only snapshot, so planning never blocks ingestion. For efficiency, the runtime fills in the plan's fixed structure itself, the model decodes only the values that require reasoning, and a live executor applies the result to the input gate, the encoder, and the cache.

We evaluate \textsc{Foresight} with a frozen Qwen3-VL-8B backbone. On OmniPro Online, \textsc{Foresight} reaches 23.0 mean F1, against 13.5 for the strongest baseline, MiniCPM-o 4.5. Figure~\ref{fig:teaser} (right) shows that this accuracy does not come at the cost of speed. \textsc{Foresight} also helps general streaming understanding, improving its backbone by 6.7 points on StreamingBench, with SOTA results, and by 15.4 points on OVO-Bench. The largest gain, 18.7 points, is on Forward Active Responding, where the answer appears only later in the video. This is the setting our argument targets: the right move is to hold the question and check again once the evidence arrives. Our contributions are:
\begin{itemize}
    \item We formulate \textbf{anticipatory computation} for streaming VLMs, in which the model uses its current understanding of the scene to decide when to reason next, what to check then, and how densely to sample until then.
    \item We introduce \textbf{\textsc{Foresight}}, a training-free architecture in which one frozen VLM perceives and plans over a shared causal state, so that planning runs asynchronously without pausing perception, and plans are decoded and executed with little overhead.
    \item We show that this inference-time controller leads all compared systems on proactive responding, improves streaming understanding, and helps most where evidence arrives later, supporting anticipation as a way to control computation, not just prediction.
\end{itemize}

%
\newcommand{\cmark}{\ensuremath{\checkmark}}
\newcommand{\pmark}{\ensuremath{\triangle}}
\newcommand{\xmark}{--}

\section{Related Work}
\label{sec:related-work}

\textbf{Streaming and proactive vision-language models.}
Early streaming VLMs interleaved video frames with text and learned when to
stay silent and when to speak \citep{chen2024videollmonline,wang2025videollm}.
Subsequent systems strengthened this streaming interaction loop in
complementary ways. MMDuet2 \citep{wang2026mmduet2} improves the response-timing
decision through supervised fine-tuning followed by reinforcement learning,
whereas StreamChat \citep{liu2024streamchat} refreshes the visual context at
every decoding step to keep generation grounded in the evolving stream.
Later systems became proactive in egocentric and live video
\citep{zhang2025proactive,zhang2025eyeswideopen,yang2025livestar,yang2026livestarpro,yan2026proactvl},
always-on systems were built around them \citep{lu2026aura,yao2026streamingharness},
and MiniCPM-o~4.5 \citep{cui2026minicpmo45realtimefullduplex} now handles
vision, audio, and speech in full duplex. To find whether model has seen enough to answer, LiveStar \citep{yang2025livestar} decodes
again when new frames stop matching its current description. Dispider \citep{qian2025dispider} trains a lightweight classifier to give the decision whether to respond or not. StreamReady
\citep{azad2026streamready} goes a step further and learns a readiness token
that is penalized for answering too early or too late. QueryStream
\citep{zhang2026querystream} uses the similarity score of query and videostream to make triggering decision without training. \\
\textbf{Asynchronous Computation.}
When the model is giving response to the user, it should be also asynchronously watching the video stream for real time computer-human interactions. Further running a large VLM on every frame to find if enough information to answer a query has been seen is expensive, so several systems separate
cheap proactivity decision from costly response generation. In Dispider \citep{qian2025dispider},
a lightweight trained classifier watches the stream and decides when to respond while a
larger model answers asynchronously. StreamBridge \citep{wang2025streambridge}
and StreamMind \citep{ding2025streammind} call the language model only when a
small trained gate fires. LION-FS \citep{li2025lionfs} pairs a fast path that
decides frame by frame whether to respond with a slow path that adds detail
once it does. Reasoning has received similar attention. VST
\citep{guan2026videostreamingthinking} spreads reasoning across playback,
R3-Streaming \citep{liu2026r3streaming} learns when to route a query to a
stronger model, and Asynchronous Reasoning \citep{yakushev2025asynchronous}
lets a model keep reading input while it thinks, without extra training. Like
adaptive computation and dynamic networks more broadly
\citep{graves2016act,huang2018msdnet,han2022dynamic,raposo2024mixtureofdepths},
these systems react to the present and decide how much
compute should it perform on input that has just arrived. Unlike other asynchronous systems that use different LLMs or encoders which have to duplicate the task of video encoding and processing, \textsc{Foresight} uses asynchronous twin VLMs with shared KV-cache. \\
\textbf{Anticipatory Computation}
Using anticipation to steer computation is much less explored. Action
anticipation has its own models and benchmarks
\citep{girdhar2021avt,grauman2022ego4d,zhao2024antgpt}, but there the
prediction is the final output. To our knowledge, StreamAgent
\citep{yang2025streamagent} is the only proactive streaming method that feeds its
predictions back into computation. It predicts
when and where task-relevant events will appear so that perception can be
pointed at them. Its planning, however, runs on a fixed schedule: every clip
triggers a caption, a set of candidate plans to be scored by unreliable LLM judge, and a trigger check.
In \textsc{Foresight}, one plan is generated by the VLM that decides when the next reasoning step happens, carries
the question that step should resolve, and sets how densely to sample in
between. \\
\textbf{Pruning.}
Another way to lower per-step cost is to spend fewer tokens on each frame.
FastV \citep{chen2024fastv} drops visual tokens that get little attention in
the early LLM layers, and DyCoke and PruneVid
\citep{tao2025dycoke,huang2025prunevid} merge tokens repeated across frames.
Streaming systems do the same online, keeping what changed
\citep{yao2025timechatonline} or what matches the query
\citep{zhang2026querystream}. CoPE-VideoLM \citep{sarkar2026copevideolm} skips
most of the encoding instead: its delta encoder turns the codec's motion
vectors and residuals into eight tokens per P-frame, so only keyframes go
through the full image encoder. All of these trim frames that have already
arrived. VisionZip \citep{yang2026visionziplongerbetternecessary} uses video context only to prune tokens while QueryStream \citep{zhang2026querystream} also utilizes query for token pruning. StreamAgent \citep{yang2025streamagent} proposes giving model agentic tool calling for pruning. \textsc{Foresight} decides beforehand how densely to sample the frames still
to come and then uses the model's recommendation of which tokens should be pruned which is much more informed about the task and activities happening in the video. \\
\textbf{Self-programming and Reconfiguration.}
A separate line of work lets a model write the procedure that runs it. In
visual programming, a language model turns each query into a short program
that calls vision modules \citep{gupta2023visprog,suris2023vipergpt}. Agents
interleave reasoning with tool calls whose results change what the model sees
next \citep{yao2023react,schick2023toolformer}, and VideoAgent
\citep{wang2024videoagent} uses this loop to pick which frames of a long video
to inspect. Other systems improve from feedback, refining their own outputs,
prompts, or pipelines
\citep{shinn2023reflexion,madaan2023selfrefine,wang2024voyager,khattab2024dspy},
and some write and rewrite agent code
\citep{zelikman2024stop,hu2025adas,yin2025godelagent}. All of them
reconfigure over input that stays fixed while they think. \textsc{Foresight} applies a
lightweight form of the same idea to a video stream where the rate of change of activities dictates how it should be processed.

%
%

\section{Foresight}
\label{sec:method}
\begin{figure}[t]
    \centering
    \includegraphics[
        width=1.07\linewidth,
        trim=100 17 9 14,
        clip
    ]{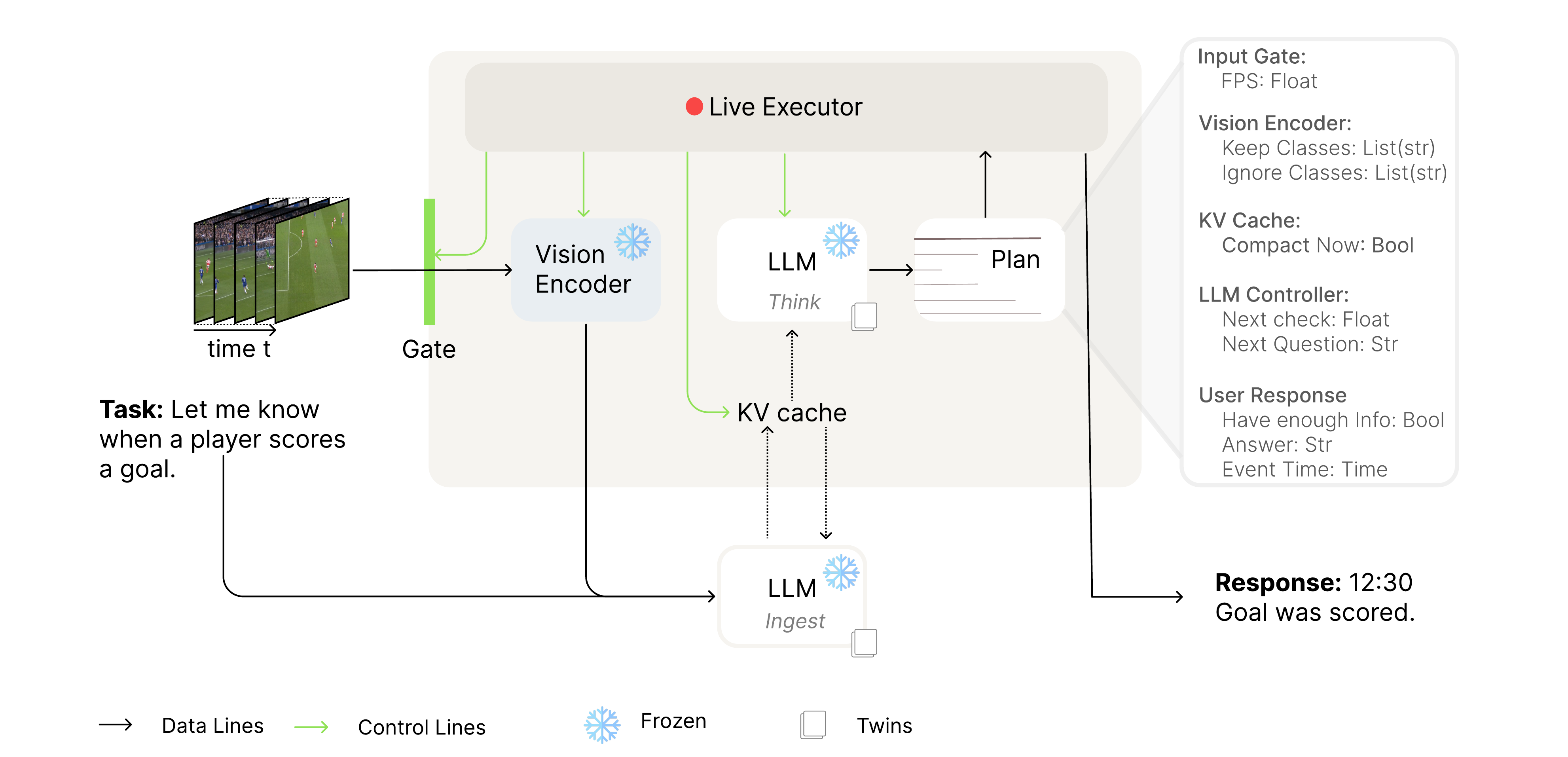}
    
    \caption{\textbf{\textsc{Foresight} overview.} One frozen VLM runs as two twins over a shared KV cache $C_t$. The \emph{Ingest LLM} (perception) appends every frame admitted by the Vision Gate to $C_t$ and never stops. When a check is due, the \emph{Think LLM} (planning) reads a snapshot of $C_t$ and writes a plan $\pi_t = (s_t, u_t)$ (right panel): an evidence state $s_t$ and a control update $u_t$. The \emph{Live Executor} applies $u_t$ along the green control lines and emits the answer once the evidence is sufficient.}

    \label{fig:foresight-overview}
\end{figure}

\begin{figure}[t]
    \centering
    \includegraphics[
        width=0.96\textwidth,
        trim=3mm 5mm 3mm 5mm,
        clip
    ]{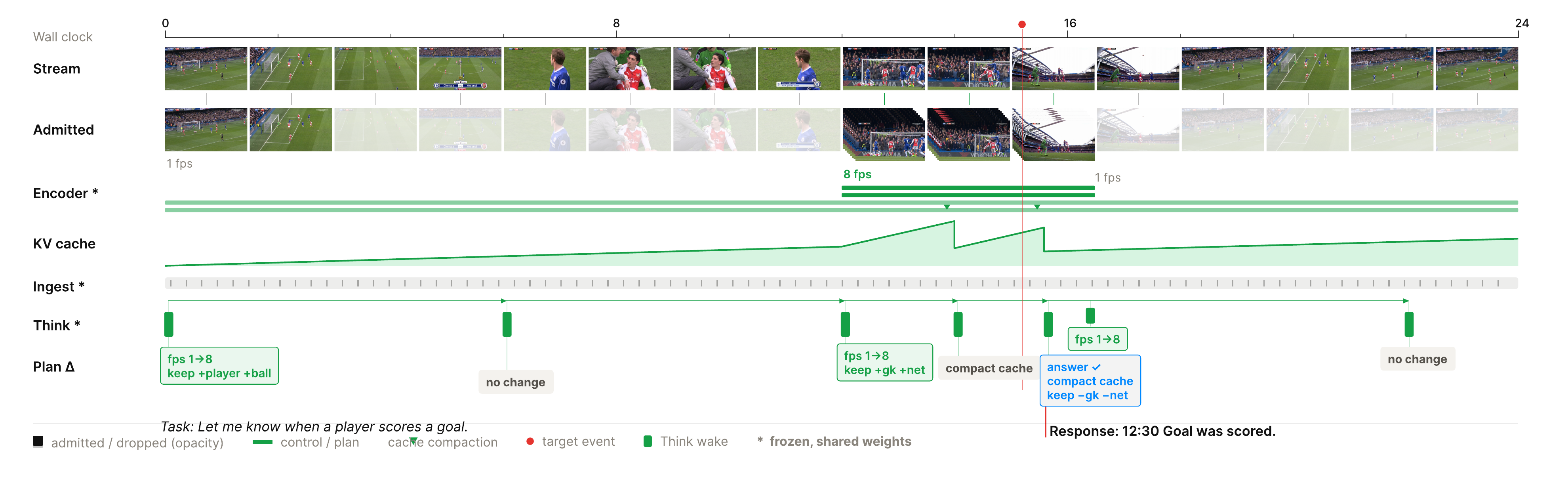}

    \caption{\textbf{\textsc{Foresight} on one stream.} Task: ``Let me know when a player scores a goal.'' Time runs left to right, and the red line marks the goal. The \emph{Ingest} never pauses (Alg.~\ref{alg:foresight}, lines 3--6), and the cache grows slowly at 1\,fps and faster at 8\,fps. The \emph{Think} LLM wakes only at scheduled checks (green bars; line 8), and the interval $\Delta$ is set by the previous plan (line 17). Early plans track only the player and ball. As a shot develops, a plan raises sampling to 8\,fps and adds the goalkeeper and net; later plans compact the cache (thus KV-cache drops). Just after the goal, the evidence is sufficient and the answer is emitted; the plan then drops the goalkeeper and net, and sampling returns to 1\,fps.}
    \label{fig:dynamic_behavior}
\end{figure}

Foresight is a
training-free method that equips an off-the-shelf VLM for proactive streaming
tasks. It combines asynchronous perception, planning, and response generation,
dynamically reconfiguring computation according to the evolving activity in
the stream to support real-time human--computer interaction. Here,
Section~\ref{sec:architecture} explains the architecture illustrated in
Figure~\ref{fig:foresight-overview}, while
Section~\ref{sec:dynamic_behavior} describes the system's dynamic behavior
illustrated in Figure~\ref{fig:dynamic_behavior} and
Algorithm~\ref{alg:foresight}.

\subsection{Architectural Overview}
\label{sec:architecture}

\textbf{Live Executor.}
As shown in Figure~\ref{fig:foresight-overview} the Live Executor takes the plan and reconfigures the system via python program. It maintains the memory of what event happened at what time.

\textbf{Vision Gate.}
It samples the stream at the frame rate set by
the plan: sparsely while nothing relevant is happening, and more densely
when the planner expects an event, so that the moments that matter are captured. Frames that are not admitted are never encoded,
which saves computation during quiet stretches of the video and intervals irrelevant to the task at hand.

\textbf{Vision Encoder.} The Vision Encoder is the VLM's own frozen image encoder that turns each admitted frame into visual tokens and then takes \texttt{keep class} and \texttt{ignore class} to dynamically prune tokens of irrelevant parts of the video as configured by the Live Executor via the generated plan.

\textbf{Ingest LLM.} The Ingest LLM writes the task query and then the visual tokens of each admitted frame into the shared KV cache $C_t$, in the order they arrive. It runs independently of the Think LLM and never waits for a plan, so perception continues while planning happens. It is also the only component that writes to $C_t$, which keeps the cache in one consistent temporal order.


\textbf{Shared KV-cache.}
It acts as the communication channel between the continuously running Ingest LLM and Think LLM so that the Think LLM can simply take the context accumulated so far and quickly generate a plan instead of encoding the inputs. To stay
within a fixed memory budget, old entries are evicted with a sliding
window, or compacted when the plan asks for it.

\textbf{Think LLM.} The Think LLM acts as the controller of \textsc{Foresight}. It reads the shared KV cache $C_t$ and writes a plan $\pi_t$ that reconfigures the system for the stream ahead. The Live Executor wakes the Think LLM at the time the last plan chose. It also passes along the question that plan left open, so the Think LLM knows what to look for. The Think LLM reads a copy of $C_t$ and never writes to it. This keeps its own reasoning out of the visual history. Instead, the Live Executor keeps a record of what it has already observed and reported (\texttt{seen} and $R_t$) and passes it to later plans.

\textbf{Plan.}
Each time the Think LLM runs, it writes a plan
\begin{equation}
    \pi_t = \bigl(s_t,u_t\bigr).
    \label{eq:plan_decomposition}
\end{equation}

\emph{Control Fields} ($u_t$). \texttt{FPS} field configures Vision Gate to sample frames at adaptive frame rate while \texttt{keep class} and \texttt{ignore class} configure which objects in the stream should get attention for performing the task. \texttt{Compact Now} sends signal to KV-cache to compact the context.\texttt{Next Check} and \texttt{Next Question} configure Think LLM by controlling when it is invoked next for updating plan and what it should pay attention to then respectively.

\emph{User Response} ($s_t$). \texttt{Have enough info} says whether the
evidence is sufficient to proactively give response. If it is, \texttt{Event time} gives when the event
happened and \texttt{Answer} to the user. 

\subsection{Dynamic Behaviour}
\label{sec:dynamic_behavior}
Figure~\ref{fig:dynamic_behavior} traces \textsc{Foresight} on one stream. The Vision Gate admits frames at the current frame rate, and the Vision Encoder turns them into visual tokens, dropping the tokens of object classes the plan ignores. The Ingest LLM appends the task query and these tokens to the shared KV cache $C_t$ without pausing. Meanwhile, at each scheduled check, the Think LLM reads a snapshot of $C_t$ and writes a plan $\pi_t$. The Live Executor then applies the plan. It compacts the cache if \texttt{Compact Now} is set, and it passes any new \texttt{FPS}, \texttt{Keep Classes}, or \texttt{Ignore Classes} to the Vision Gate and Vision Encoder, which adjust frame- and token-level pruning for the frames to come.

Algorithm~\ref{alg:foresight} gives the same procedure for a frozen VLM $\mathcal{M}$, a visual stream $V$, and a task instruction $I$. Two loops run concurrently. The perception loop (lines 3--6) encodes each admitted frame $x_t$ with $\mathcal{M}_{\mathrm{enc}}$ and appends it to the causal state, turning $C_{t^-}$ into $C_t$. The planning loop (lines 7--18) waits until video time reaches the scheduled check $\tau_{\mathrm{next}}$, takes a read-only snapshot $C_t^{\mathrm{snap}}$, and plans from it together with the open question $U.q$ and the response history $R_t$. It merges the control update into $U$ and applies it. If the output gate lets the answer through, the answer is sent to the user and added to $R_t$. Finally, the next check is set to $t + U.\Delta$. The controller starts from default $U_0$ and $\Delta_0$.

\begin{algorithm}[t]
\caption{\textsc{Foresight}: anticipatory control for streaming VLM inference}
\label{alg:foresight}
\small
\begin{algorithmic}[1]

\Require frozen VLM $\mathcal{M}$, visual stream $V$, task instruction $I$

\State $C_0 \gets \textsc{InitializeContext}(\mathcal{M},I)$
\State $U \gets U_0$; \;
       $R_0 \gets \varnothing$; \;
       $\tau_{\mathrm{next}} \gets \Delta_0$

\Statex
\Statex \textbf{The following two roles run concurrently:}

\Statex \textit{Streaming perception --- sole writer of $C_t$}

\For{each admitted observation $(t,x_t)$}

    \State $C_t
           \gets
           \textsc{Append}\!\left(
               C_{t^-},
               \mathcal{M}_{\mathrm{enc}}(x_t)
           \right)$

    \State \textsc{Publish}$(t)$

\EndFor

\Statex
\Statex \textit{Anticipatory planning --- snapshot reader}

\While{stream is active}

    \State \textbf{wait until} published video time
           $t \geq \tau_{\mathrm{next}}$

    \State $C_t^{\mathrm{snap}}
           \gets
           \textsc{Snapshot}(C_t)$

    \State $\pi_t=(s_t,u_t)
           \gets
           \textsc{Plan}\!\left(
               \mathcal{M},
               C_t^{\mathrm{snap}},
               U.q,
               R_t
           \right)$
           \Comment{Alg.~\ref{alg:schema_walk}}

    \State $U
           \gets
           \textsc{Merge}(U,u_t)$

    \State \textsc{ApplySamplingControl}$(U.\mathrm{fps})$

    \If{\textsc{ShouldEmit}$(s_t,p_{\mathrm{hit},t},t)$}
           \Comment{Alg.~\ref{alg:output_gate}}

        \State \textsc{Emit}$(s_t.\mathrm{answer})$

        \State $R_t
               \gets
               \textsc{UpdateHistory}
               (R_t,t,s_t.\mathrm{answer})$

    \EndIf

    \State $\tau_{\mathrm{next}}
           \gets
           t + U.\Delta$

\EndWhile

\end{algorithmic}
\end{algorithm}

\section{Experiments}
\label{sec:experiments}

\paragraph{Evaluation objectives.}
We evaluate whether \textsc{Foresight} can turn a frozen VLM into an effective
proactive streaming system without additional training. The experiments test
autonomous response timing, general streaming understanding, reasoning when
evidence arrives later, and the contribution of the controller and input
processing. Together, they assess both response quality and the effectiveness
of anticipation as a mechanism for controlling future inference.

\subsection{Implementation Details}
\label{sec:experimental_setup}

\textbf{Model.}
\textsc{Foresight} wraps a single off-the-shelf VLM,
Qwen3-VL-8B~\citep{bai2025qwen3vl}, whose weights stay frozen for every experiment.

Rather than feeding individual frames, wWe use Qwen3-VL's standard video-input pipeline. Consecutive admitted frames
are grouped into clips of two frames, resized using the model's native video
preprocessing, and passed to its vision encoder as video inputs rather than as
independent images. The resulting visual tokens are ingested with their video
timestamps, while the system instruction, task specification, and other text
inputs are tokenized and supplied directly to the language model. The system
instruction and task specification are prefilled once at the start of each
stream. We compare this two-frame video input with individually encoded image
frames in the input-processing ablation in Table~\ref{tab:plan_ablation}.

Plan fields persist in a standing controller state $U$, updated by
\begin{equation}
    U \leftarrow \operatorname{merge}(U,u_t).
    \label{eq:control_merge}
\end{equation}
Here, $u_t$ contains only the plan fields generated or changed at the current
step. The merge overwrites the corresponding fields in $U$ and retains the
previous values of all omitted fields. A setting such as the frame rate or the
open question therefore remains active until a later plan explicitly changes
it. Evidence and response fields are not persisted through this merge; they
are recomputed from the current stream state at each planning step. How the
plan is decoded and the thresholds are fitted is described below.

\textbf{Plan Decoding.}
Configuration fields of the plan, such as when to check next and which
question to revisit, are decoded autoregressively. Boolean decisions are
not decoded as text: at the field's position we read the logits of the
\texttt{yes} and \texttt{no} tokens and use their relative ratio,
$p = e^{z_{\text{yes}}}/(e^{z_{\text{yes}}} + e^{z_{\text{no}}})$, as a
probe that triggers a response when an event of relevance occurs. The
first planning step decodes the full plan; each later step decodes only
a diff against the current plan, rewriting the fields that change and
keeping the rest, which keeps plan updates light throughout the stream.

\textbf{Live Executor.}
The live executor takes each decoded plan and reconfigures the system
at that moment. Its execution code receives the plan values as
arguments, for example when the planner should wake next, which question
it should revisit, and, when adaptive sampling is enabled, how densely
the stream is sampled, and applies
them to the running pipeline. The configuration is therefore dynamic: it
changes as the video progresses and each new plan arrives.

\textbf{Pruning.}
The plan can also tell the vision encoder what to look at through the
\texttt{Keep Classes} and \texttt{Ignore Classes} fields generated by the Think
LLM controller. Because visual tokens do not carry explicit class labels, the
runtime embeds the controller-provided class names using the same frozen model
and compares them with the visual-token representations. These similarity
scores estimate the relevance of each token to the requested classes. Tokens
associated with ignored classes are removed, while tokens associated with kept
classes are retained before the visual sequence enters the KV cache. This
keeps the cached context focused on task-relevant evidence without introducing
an external detector or training an additional pruning module.

\textbf{Proactive Response Triggering and Threshold Fitting.}
Prompts and decoding settings are fixed within each benchmark
(Appendix~\ref{app:prompts_thresholds}). The only values fitted to data
are the thresholds that decide when to respond. We split each benchmark
at random into 20\% for calibration and 80\% for evaluation, and on the
calibration part we fit, per task, the evidence threshold
$\theta_{\text{task}}$, the gate's firing and re-arming thresholds
$\theta_{\text{hi}} > \theta_{\text{lo}}$, and a refractory interval
between responses. We also pick per task whether the gate fires every
time the evidence is above threshold or only when it first crosses it.
To avoid reporting the same event twice, a new response is dropped if
the event time the model predicts falls within the refractory interval
of an earlier one. All \textsc{Foresight} results are on the held-out
80\%; baseline numbers are taken from the full evaluation sets.

\textbf{Compute.}
All experiments are run on four NVIDIA GH200 Grace Hopper GPUs.

\subsection{Benchmarks}
We evaluate \textsc{Foresight} on three streaming benchmarks: OmniPro
\citep{zhao2026omnipro} for proactive responding, and StreamingBench
\citep{lin2024streaming} and OVO-Bench \citep{niu2025ovobench} for
general streaming video understanding.


\textbf{OmniPro Online Mode.}
In the Online mode, the task is given once at the start and the system
must decide for itself when to respond. A response counts only if it is
correct and arrives within $\pm 3$\,s of the annotated event, and we
report joint F1 over nine task types under the official protocol
(Table~\ref{tab:omnipro_online}). Since our backbone takes video only,
we use the subset of OmniPro that does not need audio. \textsc{Foresight}
reaches a mean timing F1 of 43.7 and a content accuracy of 54.6 on its
on-time responses; the per-task split is in
Appendix~\ref{app:omnipro_online_breakdown}, and
Figure~\ref{fig:teaser} (right) plots joint F1 against the real-time
factor (compute time per second of video).

\textbf{StreamingBench. }
\label{sec:streamingbench}
StreamingBench tests whether the same controller helps beyond proactive
tasks. Its 18 tasks fall into three groups: what is on screen now
(Real-Time Visual Understanding, RTVU), how sound and vision fit
together (Omni-Source), and what depends on earlier parts of the stream
(Contextual). Table~\ref{tab:streamingbench_full} summarizes the results,
and per-task results are in Appendix~\ref{app:streamingbench_breakdown}.

\textbf{OVO-Benc.} OVO-Bench groups its twelve tasks by where the evidence lies: on screen
now (Real-Time), earlier in the video (Backward), or still to come
(Forward). Table~\ref{tab:ovo_full} summarizes the results. Pleaes, find the details per-task in Appendix~\ref{app:ovo_breakdown}.

\textbf{Ablations.}
Table~\ref{tab:plan_ablation} studies the effect of the control interface.
Each row changes one part of the system while retaining the same frozen
backbone, prompts, and evaluation protocol. The field ablations isolate the
roles of scheduling, visual admission, token selection, and cache management.
The input-processing ablation instead replaces two-frame video clips with
individually encoded frames. Processing short clips allows Qwen3-VL to use its
native video pathway and preserve local temporal context between consecutive
frames. The comparison therefore separates the effect of input representation
from the effect of the controller itself.




\begin{table}[H]
\caption{\textbf{OmniPro Online Mode, joint F1 (\%)} on the subset that does not need audio. Bold is best, underline second best. The propsoed \textsc{Foresight} performs the best, despite being training-free.}
\label{tab:omnipro_online}
\centering
\scriptsize
\setlength{\tabcolsep}{3.2pt}
\renewcommand{\arraystretch}{1.12}
\begin{tabular}{llcccccccccc}
\toprule
& & \multicolumn{4}{c}{\textbf{Perception}} &
\multicolumn{3}{c}{\textbf{Comprehension}} &
\multicolumn{2}{c}{\textbf{Reasoning}} & \\
\cmidrule(lr){3-6} \cmidrule(lr){7-9} \cmidrule(lr){10-11}
\multirow{2}{*}{\textbf{Model}} & \multirow{2}{*}{\textbf{Params}} &
Event- & Target- & State- & Snap.- & Cond.- & Cum.- & Event- &
Dedup.- & Step- & \multirow{2}{*}{\textbf{Mean}} \\
& & Alert & Ground & Monitor & Count & Alert & Count & Narr. &
Count & Inst. & \\
\midrule
LiveStar~\citep{yang2025livestar} & 8B & 9.7 & 0.8 & 0.0 & 0.0 & 14.7 & 0.0 & 1.6 & 0.0 & 6.0 & 3.6 \\
MMDuet2~\citep{wang2026mmduet2} & 3B & 12.5 & 5.3 & \underline{14.9} & \underline{11.2} & 21.4 & 5.3 & 3.7 & \underline{12.7} & \underline{14.7} & 11.3 \\
MiniCPM-o 4.5~\citep{cui2026minicpmo45realtimefullduplex} & 9B & \textbf{50.1} & \underline{12.6} & 9.8 & 8.0 & \underline{23.8} & \underline{9.0} & 3.0 & 3.8 & 1.4 & \underline{13.5} \\
Dispider$^\dagger$~\citep{qian2025dispider} & 7B & 2.3 & 0.0 & 0.0 & 0.0 & 8.5 & 0.3 & 4.5 & 0.1 & 9.7 & 2.8 \\
StreamAgent~\citep{yang2025streamagent} & 8B & 6.8 & 1.4 & 4.0 & 1.9 & 11.3 & 0.9 & 3.7 & 8.1 & 7.1 & 5.0 \\
QueryStream~\citep{zhang2026querystream} & 8B & 8.2 & 1.4 & 3.1 & 1.0 & 17.6 & 6.6 & \underline{7.2} & 7.1 & 8.3 & 6.7 \\
\midrule
\textsc{Foresight} & 8B & \underline{44.7} & \textbf{17.9} & \textbf{16.4} & \textbf{19.1} & \textbf{33.8} & \textbf{16.3} & \textbf{11.5} & \textbf{29.2} & \textbf{18.5} & \textbf{23.0} \\
\bottomrule
\end{tabular}
\end{table}


\vspace{-3mm}
\begin{table}[H]
\caption{\textbf{StreamingBench, accuracy (\%).} Bold is best, underline second best.}
\label{tab:streamingbench_full}
\centering
\footnotesize
\setlength{\tabcolsep}{1.4pt}
\renewcommand{\arraystretch}{1.12}
\begin{tabular}{lccccc}
\toprule
\textbf{Method} & \textbf{FPS} & \textbf{RTVU $\uparrow$} &
\textbf{Omni-Source $\uparrow$} & \textbf{Contextual $\uparrow$} &
\textbf{Overall $\uparrow$} \\
\midrule
VideoLLM-Online~\citep{chen2024videollmonline} & 2 & 35.99 & 28.45 & 26.55 & 32.48 \\
Flash-VStream~\citep{zhang2024flashvstream} & 1 & 23.23 & 26.00 & 24.12 & 24.04 \\
Dispider~\citep{qian2025dispider} & 1 & 67.63 & 35.66 & 33.61 & 53.12 \\
TimeChat-Online$^\ddagger$~\citep{yao2025timechatonline} & 1 & 75.36 & 36.96 & 35.63 & 58.11 \\
StreamAgent~\citep{yang2025streamagent} & 1 & 74.28 & 36.26 & 34.62 & 57.02 \\
Qwen3-VL-8B-Instruct$^\ast$~\citep{bai2025qwen3vl} & -- & 74.10 & 41.90 & 39.80 & 59.30 \\
StreamBridge~\citep{wang2025streambridge}, LLaVA-OV & 1 & 68.39 & 26.60 & 31.74 & 50.96 \\
\quad LLaVA-OV + Stream-IT & 1 & 70.92 & 26.30 & 29.21 & 51.73 \\
StreamBridge~\citep{wang2025streambridge}, Qwen2-VL & 1 & 72.01 & 31.30 & 36.57 & 55.09 \\
\quad Qwen2-VL + Stream-IT & 1 & 77.04 & 24.10 & 32.55 & 55.39 \\
StreamForest$^\ddagger$~\citep{zeng2025streamforest} & 1 & 77.14 & 38.05 & 36.34 & 60.80 \\
MiniCPM-o 4.5~\citep{cui2026minicpmo45realtimefullduplex} & -- & \underline{78.20} & \underline{42.10} & \textbf{44.50} & \underline{62.70} \\
\midrule
\textsc{Foresight} & adaptive & \textbf{79.30} & \textbf{48.70} & \underline{43.80} & \textbf{66.00} \\
\bottomrule
\end{tabular}
\end{table}



\begin{table}[H]
\caption{\textbf{OVO-Bench, accuracy (\%).} Bold is best, underline second best.}
\label{tab:ovo_full}
\centering
\scriptsize
\setlength{\tabcolsep}{3.5pt}
\renewcommand{\arraystretch}{1.12}
\begin{tabular}{lccccc}
\toprule
\textbf{Method} & \textbf{Frames} & \textbf{Real-Time $\uparrow$} &
\textbf{Backward $\uparrow$} & \textbf{Forward $\uparrow$} &
\textbf{Overall $\uparrow$} \\
\midrule
Human & native & 93.20 & 92.30 & 92.90 & 92.80 \\
Flash-VStream~\citep{zhang2024flashvstream} & 1 fps & 28.37 & 27.38 & 45.09 & 33.61 \\
Dispider~\citep{qian2025dispider} & 1 fps & 54.55 & 36.06 & 34.72 & 41.78 \\
TimeChat-Online~\citep{yao2025timechatonline} & 1 fps & 58.60 & 42.00 & 36.40 & 45.60 \\
StreamForest~\citep{zeng2025streamforest} & 1 fps & 61.20 & 52.02 & 53.49 & 55.57 \\
QueryStream~\citep{zhang2026querystream} & -- & 61.40 & 42.10 & 39.03 & 47.51 \\
StreamAgent~\citep{yang2025streamagent} & 1 fps & 61.30 & 41.70 & 45.40 & 49.40 \\
Qwen3-VL-8B-Instruct$^\ast$~\citep{bai2025qwen3vl} & -- & 61.60 & 41.00 & 37.70 & 46.80 \\
MiniCPM-o 4.5~\citep{cui2026minicpmo45realtimefullduplex} & 1 fps & 67.30 & \underline{55.90} & \underline{55.90} & 59.70 \\
StreamBridge~\citep{wang2025streambridge}, Qwen2-VL-7B + Stream-IT & 1 fps & \underline{71.30} & \textbf{68.05} & 48.36 & \textbf{62.57} \\
\midrule
\textsc{Foresight} & adaptive & \textbf{75.02} & 55.10 & \textbf{56.37} & \underline{62.16} \\
\bottomrule
\end{tabular}
\end{table}


\begin{table}[H]
\caption{\textbf{Controller and input-processing ablations on OmniPro Online
(\%).} Plan-field rows remove the named control while keeping everything else
fixed. The input-processing row replaces two-frame clips with individual-frame
inputs.}
\label{tab:plan_ablation}
\centering
\scriptsize
\setlength{\tabcolsep}{10pt}
\renewcommand{\arraystretch}{1.10}
\begin{tabular}{llcc}
\toprule
\textbf{Component} & \textbf{Ablation} & \textbf{Time F1} &
\textbf{Joint F1} \\
\midrule
\multicolumn{2}{l}{Full \textsc{Foresight} (as reported)} & 43.7 & 23.0 \\
\midrule
Think LLM & Next Check & 43.0 & 18.9 \\
Think LLM & Next Question & 43.3 & 19.4 \\
Think LLM & Next Check + Next Question & 43.2 & 18.6 \\
Vision Gate & FPS & 43.5 & 23.0 \\
Vision Encoder & Keep Classes & 42.7 & 22.7 \\
Vision Encoder & Ignore Classes & 43.5 & 22.9 \\
KV Cache & Compact Now & 44.3 & 23.4 \\
\midrule
Live Executor & all fields (plan ignored) & 4.7 & 0.7 \\
User Response & all but Answer & 4.0 & 0.3 \\
\midrule
Input Processing & Individual frames (no two-frame clips) & 40.1 & 19.3 \\
\bottomrule
\end{tabular}
\end{table}

\section{Conclusion}
We presented \textsc{Foresight}, a training-free framework that turns a frozen, off-the-shelf VLM into a proactive and asynchronous streaming system. Our central argument is that a streaming VLM already understands a scene well enough to know what it is still missing, and can therefore plan its own future perception instead of reacting only to what has already arrived. \textsc{Foresight} realizes this idea with two copies of the same model over a shared KV cache: one keeps ingesting the stream, while the other plans in the background when to reason next, what to check then, and how densely to sample, so that planning never pauses perception. The experiments support this design. With a frozen Qwen3-VL-8B backbone, \textsc{Foresight} reaches 23.0 joint F1 on OmniPro Online, against 13.5 for the strongest trained baseline, while running close to real time. It also achieves the best overall score on StreamingBench and improves its backbone by 15.4 points on OVO-Bench, with the largest gain, 18.7 points, on Forward Active Responding, i.e. the setting where the answer appears only later in the stream. Because no weights are updated, the same controller can directly benefit from stronger backbones as they become available. More broadly, our results suggest that anticipation is most valuable not as a prediction to be scored, but as a way for a model to decide how to spend its own computation on what comes next.

\section{Limitations and Future Work}
\label{sec:limitations}

The accuracy of \textsc{Foresight}'s anticipation and temporal event
localization depends on the capacity of the underlying VLM. Because the
controller relies heavily on these abilities to decide when to reason and
respond, reliable operation requires a VLM with strong future anticipation and
temporal grounding. Future work should improve these capabilities through
stronger backbones or dedicated inference-time mechanisms. After an initial false trigger, the model can occasionally continue reporting
an event even when subsequent observations do not support it. Providing the
Think LLM with a history of what has been observed and already reported reduces
this behavior substantially, but does not eliminate it. The controller also remains sensitive to prompt wording, motivating
future work on more prompt-robust control policies.

\section{AI Use Statement}
In this work, we used generative AI tools to assist in implementing parts of
the evaluation and inference code and in correcting \LaTeX{} formatting. We
did not use generative AI tools to develop the conceptual framework, propose
hypotheses, design the methodology or experiments, or interpret results;
generating synthetic data, mathematical claims, proofs, and translation are
not applicable to this work. Separately, as part of the benchmark protocol,
GPT-4o-mini is used as an automatic judge of content correctness for
Event-Narration and Step-Instruction
(Appendix~\ref{app:omnipro_online_breakdown}). All AI-assisted code was reviewed and
tested, and we take full responsibility for the content of this
work.

\renewcommand{\refname}{References}
\bibliographystyle{plainnat}
\bibliography{iclr2027_conference}

%


\newcommand{\Plan}{\textsc{Plan}}
\newcommand{\Prefill}{\textsc{Prefill}}
\newcommand{\Decode}{\textsc{Decode}}
\newcommand{\ReadProb}{\textsc{ReadProb}}
\newcommand{\ParseTail}{\textsc{ParseTail}}
\newcommand{\ShouldEmit}{\textsc{ShouldEmit}}

\algrenewcommand\algorithmicrequire{\textbf{Input:}}
\algrenewcommand\algorithmicensure{\textbf{Output:}}


\clearpage
\appendix 
\section*{Appendix}

\section{Detailed Algorithms}
\label{app:detailed_algorithms}
\suppressfloats[t]

\setcounter{algorithm}{0}
\renewcommand{\thealgorithm}{S\arabic{algorithm}}
\renewcommand{\theHalgorithm}{S\arabic{algorithm}}


\begin{algorithm}[H]
\caption{\Plan: schema-walked generation of the control plan
$\pi_t=(s_t,u_t)$}
\label{alg:schema_walk}
\small

\begin{algorithmic}[1]

\Require causal snapshot $C_t^{\mathrm{snap}}$;
         deferred question $q$;
         emitted-response history $R_t$;
         task threshold $\theta_{\mathrm{task}}$;
         optional-control threshold $\theta_{\mathrm{more}}$

\Ensure transient evidence state $s_t$;
        future-control update $u_t$;
        evidence confidence $p_{\mathrm{hit},t}$

\State $P_t \gets \textsc{Prompt}(q,R_t)$
\State $u_t \gets \varnothing$

\Statex
\Statex \textit{Current observation grounding}

\State $z \gets
       \Prefill\!\left(
       C_t^{\mathrm{snap}},
       P_t \Vert \texttt{\{"seen":"}
       \right)$

\State $(s_t.\mathrm{seen},z)
       \gets
       \Decode(z,\text{short text})$

\Statex
\Statex \textit{Evidence sufficiency}

\State $z \gets
       \Prefill\!\left(
       z,
       \texttt{","have\_enough\_info":}
       \right)$

\State
$p_{\mathrm{hit},t}
\gets
\frac{
    \exp z(\texttt{true})
}{
    \exp z(\texttt{true})
    +
    \exp z(\texttt{false})
}$

\State $s_t.\mathrm{hit}
       \gets
       [p_{\mathrm{hit},t}
       \geq
       \theta_{\mathrm{task}}]$

\State $s_t.\mathrm{event\_time}\gets\bot$;
       \quad
       $s_t.\mathrm{answer}\gets\varnothing$

\If{$s_t.\mathrm{hit}$}

    \State $z \gets
           \Prefill\!\left(
           z,
           \texttt{true,"event\_time\_s":}
           \right)$

    \State $(s_t.\mathrm{event\_time},z)
           \gets
           \Decode(z,\text{time})$

    \State $z \gets
           \Prefill\!\left(
           z,
           \texttt{,"answer":"}
           \right)$

    \State $(s_t.\mathrm{answer},z)
           \gets
           \Decode(z,\text{short response})$

    \State $z \gets
           \Prefill\!\left(
           z,
           \texttt{","more":}
           \right)$

\Else

    \State $z \gets
           \Prefill\!\left(
           z,
           \texttt{false,"more":}
           \right)$

\EndIf

\Statex
\Statex \textit{Future-control update}

\State $p_{\mathrm{more}}
       \gets
       \ReadProb(z)$

\If{$p_{\mathrm{more}}
    \geq
    \theta_{\mathrm{more}}$}

    \State $z \gets
           \Prefill(z,\texttt{true,})$

    \State $(\mathrm{tail},z)
           \gets
           \Decode(z,\text{optional control tail})$

    \State $u_t
           \gets
           \ParseTail(\mathrm{tail})$

    \Statex
    \Comment{$u_t$ may update
    \texttt{next\_check\_s},
    \texttt{question\_for\_next},
    and \texttt{fps}}

\EndIf

\State \Return
       $(s_t,u_t,p_{\mathrm{hit},t})$

\end{algorithmic}
\end{algorithm}


\begin{algorithm}[H]
\caption{\ShouldEmit: temporally gated response decision}
\label{alg:output_gate}
\small

\begin{algorithmic}[1]

\Require transient evidence $s_t$;
         evidence confidence $p_{\mathrm{hit},t}$;
         current video time $t$;
         gate state
         $G=(a,t_{\mathrm{fire}})$;
         thresholds
         $\theta_{\mathrm{hi}},
          \theta_{\mathrm{lo}}$;
         re-arm interval $\rho$;
         debounce interval $d$

\Ensure emission decision $\mathrm{fire}$;
        updated gate state $G$

\Statex
\Statex \textit{Re-arm the temporal gate}

\If{$p_{\mathrm{hit},t}<\theta_{\mathrm{lo}}
    \ \lor\
    t-t_{\mathrm{fire}}\geq\rho$}

    \State $a\gets1$

\EndIf

\Statex
\Statex \textit{Determine whether the current evidence defines a new event}

\State $\mathrm{candidate}
       \gets
       [s_t.\mathrm{answer}\neq\varnothing]$

\State $\mathrm{fire}
       \gets
       \mathrm{candidate}
       \land a
       \land
       [p_{\mathrm{hit},t}\geq\theta_{\mathrm{hi}}]
       \land
       [t-t_{\mathrm{fire}}>d]$

\If{$\mathrm{fire}$}

    \State $a\gets0$
    \State $t_{\mathrm{fire}}\gets t$

\EndIf

\State $G\gets(a,t_{\mathrm{fire}})$

\State \Return $(\mathrm{fire},G)$

\end{algorithmic}
\end{algorithm}

\section{Experimental Implementation Details}
\label{app:experimental_details}
\suppressfloats[t]

\subsection{Compute Infrastructure}
\label{app:compute_details}
\suppressfloats[t]

All reported full-benchmark experiments were executed on NVIDIA GH200 GPUs on
an HPC cluster.

We split each benchmark's official evaluation set at random into a 20\%
calibration split, used only to fit the response thresholds
(Section~\ref{sec:experimental_setup}), and an 80\% held-out split on
which every reported \textsc{Foresight} number is computed.

Streaming evaluation is far more expensive than single-shot video
question answering, because frames are ingested causally and the
planner may run many times per video. A complete OmniPro evaluation of
one configuration took about \textbf{150 GH200 GPU-hours}, covering
ingestion, planning, response generation, and scoring.

In the reported results the visual sampling rate is set by the plan
through its \texttt{fps} field, as listed in the Frames column of each
table; the effect of removing this control is measured in
Table~\ref{tab:plan_ablation}.

\subsection{In-Context Learning and Evaluation Prompts}
\label{app:prompts_thresholds}
\suppressfloats[t]

\textsc{Foresight} trains nothing; it relies on two abilities the frozen
VLM already has. Its world knowledge tells it how scenes tend to unfold,
and in-context learning lets it follow the plan format from instructions
in the prompt alone. Three examples show how the two combine.

\begin{itemize}
    \item \emph{Sports: ``tell me when a goal is scored.''} A player
    dribbles into the box. The model knows a shot may follow within
    seconds, so it brings \texttt{next\_check\_s} forward, raises
    \texttt{fps}, and sets \texttt{question\_for\_next} to ``has the ball
    crossed the line?''.
    \item \emph{Cooking: ``tell me when the water boils.''} A pot has just
    been put on the stove. The model knows boiling takes minutes, not
    seconds, so it sets a long \texttt{next\_check\_s}, keeps
    \texttt{fps} low, and asks next time ``are bubbles rising?''.
    \item \emph{Counting: ``how many people enter the room?''} Someone
    walks towards the door. The model expects an entry soon and checks
    again shortly, and because one person has already been reported in
    $R_t$, it sets \texttt{new\_event} only when a different person
    comes in, so the same entry is not counted twice.
\end{itemize}

In each case world knowledge says \emph{when} something is likely to
happen, and in-context learning turns that into a plan. No weights
change.

For evaluation we use each benchmark's own task-specific prompt as
provided, so that answers are scored under its official protocol. The
prompt is prefilled once at the start of the stream and kept as a
persistent prefix of the KV cache, and text written by the planner is
never inserted into the visual history. We do not modify these prompts
after seeing evaluation results, and the response thresholds are fitted
only on the 20\% calibration split.


\clearpage
\section{Additional OmniPro Analysis}
\label{app:omnipro_details}
\suppressfloats[t]

\subsection{Online-Mode Task Breakdown}
\label{app:omnipro_online_breakdown}
\suppressfloats[t]

Table~\ref{tab:omnipro_online_detailed} reports task-level Time F1,
content accuracy and Joint F1. EA denotes
Event-Alert; TG, Target-Ground; SM, State-Monitor; SC, Snapshot-Count;
CA, Conditional-Alert; CC, Cumulative-Count; EN, Event-Narration; DC,
Deduplicated-Count; and SI, Step-Instruction.
Content correctness for Event-Narration and Step-Instruction is judged by
GPT-4o-mini on the benchmark's five-level rubric, where a score of 3 or more
counts as correct; all such responses were judged.

Dispider is evaluated on the 932-example audio-helpful/none subset
supplied in its report, which gives no task-level content, so its rows
are not directly comparable with the full-benchmark ones.
MiniCPM-o~4.5 is our own run of the released checkpoint; for the same
model and setting \citet{zhao2026omnipro} publish per-task Joint F1 of
14.9, 8.7, 23.3, 16.0, 15.7, 7.6, 3.5, 27.3 and 7.5, for a mean of
13.8. StreamAgent's macro means are 12.5 (Time F1), 43.0 (content) and
5.0 (Joint F1), against the pooled values shown in the table.

\begin{table}[!htbp]
\caption{\textbf{OmniPro Online, per task (\%).} The three blocks
decompose the same emissions: Time F1 counts responses that land
inside the $\pm 3$\,s window, content accuracy how many of those were
right, and Joint F1 requires both. The two alert tasks are
content-trivial once timed, so their content is 100 by construction.
Time F1 $\times$ content $=$ Joint F1 holds for each task, but each
source pools the Overall column differently, so it need not multiply
out there. Dispider's content row is recovered
from its other two rows task by task, so its Overall is their macro
mean.}
\label{tab:omnipro_online_detailed}
\centering
\scriptsize
\setlength{\tabcolsep}{3.2pt}
\renewcommand{\arraystretch}{1.12}
\begin{tabular}{lcccccccccc}
\toprule
\textbf{Method} & \textbf{EA} & \textbf{TG} &
\textbf{SM} & \textbf{SC} & \textbf{CA} & \textbf{CC} & \textbf{EN} &
\textbf{DC} & \textbf{SI} & \textbf{Overall} \\
\midrule
\multicolumn{11}{l}{\emph{Time F1}} \\
Dispider~\citep{qian2025dispider} & 2.3 & 7.7 & 8.5 & 2.7 & 8.5 & 7.3 & 9.5 & 8.8 & 15.0 & 9.6 \\
StreamAgent~\citep{yang2025streamagent} & 6.8 & 8.8 & 10.5 & 9.0 & 11.3 & 5.7 & 16.0 & 17.2 & 27.3 & 15.0 \\
QueryStream~\citep{zhang2026querystream} & 8.2 & 6.9 & 24.9 & 7.5 & 17.6 & 22.1 & 19.9 & 21.1 & 30.9 & 17.7 \\
MiniCPM-o 4.5~\citep{cui2026minicpmo45realtimefullduplex} & 50.1 & 53.9 & 20.9 & 24.2 & 23.8 & 22.5 & 8.5 & 7.7 & 8.3 & 24.4 \\
\textsc{Foresight} & 44.7 & 47.9 & 52.6 & 48.3 & 33.8 & 41.7 & 29.6 & 41.3 & 53.2 & \textbf{43.7} \\
\midrule
\multicolumn{11}{l}{\emph{Content accuracy}} \\
Dispider~\citep{qian2025dispider} & 100.0 & 0.0 & 0.0 & 0.0 & 100.0 & 4.5 & 46.9 & 1.5 & 64.9 & 35.3 \\
StreamAgent~\citep{yang2025streamagent} & 100.0 & 16.1 & 37.9 & 21.2 & 100.0 & 15.4 & 23.2 & 46.9 & 26.2 & 36.2 \\
QueryStream~\citep{zhang2026querystream} & 100.0 & 19.6 & 12.4 & 13.1 & 100.0 & 29.7 & 36.4 & 33.5 & 26.8 & 41.3 \\
MiniCPM-o 4.5~\citep{cui2026minicpmo45realtimefullduplex} & 100.0 & 22.7 & 49.1 & 33.3 & 100.0 & 39.9 & 38.3 & 50.8 & 18.2 & 54.4 \\
\textsc{Foresight} & 100.0 & 37.4 & 31.2 & 39.6 & 100.0 & 39.1 & 38.9 & 70.8 & 34.7 & \textbf{54.6} \\
\midrule
\multicolumn{11}{l}{\emph{Joint F1}} \\
LiveStar~\citep{yang2025livestar} & 9.7 & 0.8 & 0.0 & 0.0 & 14.7 & 0.0 & 1.6 & 0.0 & 6.0 & 3.6 \\
MMDuet2~\citep{wang2026mmduet2} & 12.5 & 5.3 & 14.9 & 11.2 & 21.4 & 5.3 & 3.7 & 12.7 & 14.7 & 11.3 \\
Dispider~\citep{qian2025dispider} & 2.3 & 0.0 & 0.0 & 0.0 & 8.5 & 0.3 & 4.5 & 0.1 & 9.7 & 2.8 \\
StreamAgent~\citep{yang2025streamagent} & 6.8 & 1.4 & 4.0 & 1.9 & 11.3 & 0.9 & 3.7 & 8.1 & 7.1 & 5.0 \\
QueryStream~\citep{zhang2026querystream} & 8.2 & 1.4 & 3.1 & 1.0 & 17.6 & 6.6 & 7.2 & 7.1 & 8.3 & 6.7 \\
MiniCPM-o 4.5~\citep{cui2026minicpmo45realtimefullduplex} & 50.1 & 12.6 & 9.8 & 8.0 & 23.8 & 9.0 & 3.0 & 3.8 & 1.4 & 13.5 \\
\textsc{Foresight} & 44.7 & 17.9 & 16.4 & 19.1 & 33.8 & 16.3 & 11.5 & 29.2 & 18.5 & \textbf{23.0} \\
\bottomrule
\end{tabular}
\end{table}

\subsection{Emission Behaviour of the Baselines}
\label{app:omnipro_audio_emission}
\suppressfloats[t]

How often a system speaks is worth separating from how well it is
timed. Against the 9,051 ground-truth events, StreamAgent emits
0.79 responses per event and our MiniCPM-o~4.5 run only 0.37, yet
MiniCPM-o~4.5 attains the higher Time F1 of the two (24.4 against
15.0). QueryStream sits at the opposite extreme at 2.75 responses per
event, which lifts its Time F1 to 17.7 but, with most of those
responses wrong, keeps its joint F1 low.


\clearpage
\section{Additional StreamingBench Results}
\label{app:streamingbench_extra}
\suppressfloats[t]

\subsection{Task-Level Breakdown}
\label{app:streamingbench_breakdown}
\suppressfloats[t]

We split the original 18-task matrix into two readable tables. OP denotes
Object Perception; CR, Causal Reasoning; CS, Clips Summarization; ATP,
Attribute Perception; EU, Event Understanding; TR, Text-Rich
Understanding; PR, Prospective Reasoning; SU, Spatial Understanding;
ACP, Action Perception; and CT, Counting.

\begin{table}[!htbp]
\caption{\textbf{StreamingBench Real-Time Visual Understanding, per
task (accuracy, \%).} The first two blocks are reference points rather
than streaming systems, taken from \citet{lin2024streaming}; a plain
frame count means uniform sampling rather than a streaming rate.
TimeChat-Online's keep-rate variants are from
\citet{yao2025timechatonline} and the Qwen3-VL-8B and MiniCPM-o~4.5
rows from \citet{lu2026aura}.}
\label{tab:streamingbench_rtvu_detailed}
\centering
\scriptsize
\setlength{\tabcolsep}{2.2pt}
\renewcommand{\arraystretch}{1.10}
\begin{tabular}{lcccccccccccc}
\toprule
\textbf{Method} & \textbf{FPS} & \textbf{OP} & \textbf{CR} & \textbf{CS} &
\textbf{ATP} & \textbf{EU} & \textbf{TR} & \textbf{PR} & \textbf{SU} &
\textbf{ACP} & \textbf{CT} & \textbf{Avg.} \\
\midrule
Human & native & 89.47 & 92.00 & 93.60 & 91.47 & 95.65 & 92.52 & 88.00 & 88.75 & 89.74 & 91.30 & 91.46 \\
\midrule
Gemini 1.5 Pro~\citep{team2024gemini15} & 1 & 79.02 & 80.47 & 83.54 & 79.67 & 80.00 & 84.74 & 77.78 & 64.23 & 71.95 & 48.70 & 75.69 \\
GPT-4o~\citep{openai2024gpt4o} & 64 & 77.11 & 80.47 & 83.91 & 76.47 & 70.19 & 83.80 & 66.67 & 62.19 & 69.12 & 49.22 & 73.28 \\
Claude 3.5 Sonnet~\citep{anthropic2024claude35sonnet} & 20 & 80.49 & 77.34 & 82.02 & 81.73 & 72.33 & 75.39 & 61.11 & 61.79 & 69.32 & 43.09 & 72.44 \\
\midrule
Video-LLaMA2-7B~\citep{cheng2024videollama2} & 32 & 55.86 & 55.47 & 57.41 & 58.17 & 52.80 & 43.61 & 39.81 & 42.68 & 45.61 & 35.23 & 49.52 \\
VILA-1.5-8B~\citep{lin2024vila} & 14 & 53.68 & 49.22 & 70.98 & 56.86 & 53.42 & 53.89 & 54.63 & 48.78 & 50.14 & 17.62 & 52.32 \\
Video-CCAM-14B~\citep{fei2024videoccam} & 96 & 56.40 & 57.81 & 65.30 & 62.75 & 64.60 & 51.40 & 42.59 & 47.97 & 49.58 & 31.61 & 53.96 \\
LongVA-7B~\citep{zhang2024longva} & 128 & 70.03 & 63.28 & 61.20 & 70.92 & 62.73 & 59.50 & 61.11 & 53.66 & 54.67 & 34.72 & 59.96 \\
InternVL2-8B~\citep{chen2024internvl2} & 16 & 68.12 & 60.94 & 69.40 & 77.12 & 67.70 & 62.93 & 59.26 & 53.25 & 54.96 & 56.48 & 63.72 \\
Kangaroo-7B~\citep{liu2024kangaroo} & 64 & 71.12 & 84.38 & 70.66 & 73.20 & 67.08 & 61.68 & 56.48 & 55.69 & 62.04 & 38.86 & 64.60 \\
LLaVA-NeXT-Video-32B~\citep{zhang2024llavanextvideo} & 64 & 78.20 & 70.31 & 73.82 & 76.80 & 63.35 & 69.78 & 57.41 & 56.10 & 64.31 & 38.86 & 66.96 \\
MiniCPM-V 2.6-8B~\citep{yao2024minicpmv} & 32 & 71.93 & 71.09 & 77.92 & 75.82 & 64.60 & 65.73 & 70.37 & 56.10 & 62.32 & 53.37 & 67.44 \\
LLaVA-OneVision-7B~\citep{li2024llavaonevision} & 32 & 80.38 & 74.22 & 76.03 & 80.72 & 72.67 & 71.65 & 67.59 & 65.45 & 65.72 & 45.08 & 71.12 \\
Qwen2.5-VL-7B~\citep{bai2025qwen25vl} & 1 & 78.32 & 80.47 & 78.86 & 80.45 & 76.73 & 78.50 & 79.63 & 63.41 & 66.19 & 53.19 & 73.68 \\
Qwen3-VL-8B-Instruct~\citep{bai2025qwen3vl} & -- & 80.10 & 76.60 & 81.10 & 81.10 & 77.00 & 73.20 & 80.60 & 63.40 & 68.80 & 57.00 & 74.10 \\
\midrule
VideoLLM-Online~\citep{chen2024videollmonline} & 2 & 39.07 & 40.06 & 34.49 & 31.05 & 45.96 & 32.40 & 31.48 & 34.16 & 42.49 & 27.89 & 35.99 \\
Flash-VStream~\citep{zhang2024flashvstream} & 1 & 25.89 & 43.57 & 24.91 & 23.87 & 27.33 & 13.08 & 18.52 & 25.20 & 23.87 & 48.70 & 23.23 \\
Dispider~\citep{qian2025dispider} & 1 & 74.92 & 75.53 & 74.10 & 73.08 & 74.44 & 59.92 & 76.14 & 62.91 & 62.16 & 45.80 & 67.63 \\
TimeChat-Online~\citep{yao2025timechatonline} & 1 & 80.22 & 82.03 & 79.50 & 83.33 & 76.10 & 78.50 & 78.70 & 64.63 & 69.60 & 57.98 & 75.36 \\
\quad 55.8\% keep rate & 1 & 80.76 & 79.69 & 80.76 & 83.33 & 74.84 & 78.82 & 78.70 & 64.23 & 68.75 & 57.98 & 75.28 \\
\quad 17.4\% keep rate & 1 & 79.13 & 81.25 & 78.86 & 80.77 & 70.44 & 77.26 & 77.78 & 67.07 & 66.19 & 53.72 & 73.64 \\
StreamAgent~\citep{yang2025streamagent} & 1 & 79.63 & 78.31 & 79.28 & 75.87 & 74.74 & 76.92 & 82.94 & 66.31 & 73.69 & 55.40 & 74.28 \\
QueryStream~\citep{zhang2026querystream} & 1 & 82.38 & 84.38 & 79.18 & 82.37 & 77.99 & 81.31 & 78.70 & 65.04 & 69.32 & 47.34 & 75.32 \\
StreamBridge~\citep{wang2025streambridge}, LLaVA-OV-7B & 1 & 76.84 & 77.17 & 82.60 & 75.25 & 64.15 & 64.17 & 75.00 & 61.38 & 61.19 & 46.11 & 68.39 \\
\quad LLaVA-OV-7B + Stream-IT & 1 & 82.29 & 72.44 & 92.09 & 80.86 & 71.07 & 74.46 & 75.00 & 62.20 & 70.26 & 28.50 & 70.92 \\
StreamBridge~\citep{wang2025streambridge}, Qwen2-VL-7B & 1 & 80.38 & 78.74 & 83.22 & 79.86 & 74.21 & 69.47 & 77.78 & 63.41 & 69.97 & 43.01 & 72.01 \\
\quad Qwen2-VL-7B + Stream-IT & 1 & 84.74 & 82.68 & 88.92 & 89.77 & 77.36 & 85.36 & 84.26 & 69.92 & 71.67 & 35.75 & 77.04 \\
StreamForest~\citep{zeng2025streamforest} & 1 & 83.11 & 82.81 & 82.65 & 84.26 & 77.50 & 78.19 & 76.85 & 69.11 & 75.64 & 54.40 & 77.26 \\
\midrule
MiniCPM-o 4.5-9B~\citep{cui2026minicpmo45realtimefullduplex} & -- & 80.90 & 88.30 & 82.00 & 85.00 & 78.90 & 85.40 & 78.70 & 67.10 & 75.60 & 56.00 & 78.20 \\
\midrule
\textsc{Foresight} & adaptive & 83.7 & 76.8 & 90.5 & 88.6 & 79.1 & 85.0 & 84.3 & 75.9 & 75.5 & 43.2 & 79.30 \\
\bottomrule
\end{tabular}
\end{table}

ER denotes Emotion Recognition; SCU, Scene Understanding; SD, Source
Discrimination; MA, Multimodal Alignment; ACU, Anomaly Context
Understanding; MCU, Misleading Context Understanding; SQA, Sequential
Question Answering; and PO, Proactive Output. The Contextual columns
follow the benchmark's own order, which places ACU before MCU.

\begin{table}[!htbp]
\caption{\textbf{StreamingBench Omni-Source and Contextual, per task
(accuracy, \%).} These are the systems with results on the full 18-task
evaluation; $^\ddagger$ marks our own run under the official protocol.
Baseline Contextual averages include Proactive Output, as in their
sources. For \textsc{Foresight}, the Contextual average and overall
cover the three multiple-choice tasks, since Proactive Output scores
timing rather than answer choice and is reported separately (46.3,
95\% CI 39.1--53.7; Appendix~\ref{app:streamingbench_po}); averaged over
all four tasks, its Contextual score is 44.50.}
\label{tab:streamingbench_other_detailed}
\centering
\scriptsize
\setlength{\tabcolsep}{2.8pt}
\renewcommand{\arraystretch}{1.10}
\begin{tabular}{lccccccccccc}
\toprule
& \multicolumn{5}{c}{\textbf{Omni-Source}} &
\multicolumn{5}{c}{\textbf{Contextual}} & \\
\cmidrule(lr){2-6}\cmidrule(lr){7-11}
\textbf{Method} & \textbf{ER} & \textbf{SCU} & \textbf{SD} & \textbf{MA} &
\textbf{Avg.} & \textbf{ACU} & \textbf{MCU} & \textbf{SQA} & \textbf{PO} &
\textbf{Avg.} & \textbf{Overall} \\
\midrule
Flash-VStream~\citep{zhang2024flashvstream} & 25.91 & 24.90 & 25.60 & 28.40 & 26.00 & 24.80 & 25.20 & 26.80 & 1.96 & 24.12 & 24.04 \\
VideoLLM-Online~\citep{chen2024videollmonline} & 31.20 & 26.51 & 24.10 & 32.00 & 28.45 & 24.19 & 29.20 & 30.80 & 3.92 & 26.55 & 32.48 \\
Dispider~\citep{qian2025dispider} & 35.46 & 25.26 & 38.57 & 43.34 & 35.66 & 39.62 & 27.65 & 34.80 & 25.34 & 33.61 & 53.12 \\
StreamAgent~\citep{yang2025streamagent} & 35.86 & 26.26 & 38.87 & 44.04 & 36.26 & 39.72 & 30.25 & 39.60 & 28.90 & 34.62 & 57.02 \\
TimeChat-Online$^\ddagger$~\citep{yao2025timechatonline} & 36.62 & 26.75 & 38.55 & 45.90 & 36.96 & 41.45 & 32.27 & 40.55 & 28.25 & 35.63 & 58.11 \\
StreamForest$^\ddagger$~\citep{zeng2025streamforest} & 37.71 & 27.23 & 39.95 & 47.30 & 38.05 & 42.33 & 33.38 & 40.14 & 29.50 & 36.34 & 60.80 \\
StreamBridge~\citep{wang2025streambridge}, LLaVA-OV-7B & 24.40 & 12.00 & 32.40 & 37.60 & 26.60 & 20.00 & 19.60 & 34.40 & 52.94 & 31.74 & 50.96 \\
\quad LLaVA-OV-7B + Stream-IT & 20.80 & 13.20 & 43.60 & 27.60 & 26.30 & 20.40 & 17.60 & 41.60 & 37.26 & 29.21 & 51.73 \\
StreamBridge~\citep{wang2025streambridge}, Qwen2-VL-7B & 32.00 & 15.20 & 39.60 & 38.40 & 31.30 & 25.20 & 21.20 & 33.20 & 66.67 & 36.57 & 55.09 \\
\quad Qwen2-VL-7B + Stream-IT & 18.00 & 13.20 & 43.60 & 21.60 & 24.10 & 14.00 & 17.20 & 48.00 & 50.98 & 32.55 & 55.39 \\
Qwen3-VL-8B-Instruct~\citep{bai2025qwen3vl} & 44.40 & 27.20 & 46.40 & 49.60 & 41.90 & 30.00 & 38.40 & 38.00 & 52.80 & 39.80 & 59.30 \\
MiniCPM-o 4.5~\citep{cui2026minicpmo45realtimefullduplex} & 44.40 & 28.00 & 41.60 & 54.40 & 42.10 & 35.20 & 44.00 & 51.20 & 47.60 & 44.50 & 62.70 \\
\midrule
\textsc{Foresight} & 42.2 & 29.7 & 53.0 & 69.5 & 48.70 & 46.8 & 33.2 & 51.7 & 46.3 & 43.80 & 66.00 \\
\bottomrule
\end{tabular}
\end{table}

\subsection{Proactive Output}
\label{app:streamingbench_po}
\suppressfloats[t]

Proactive Output is the one StreamingBench task that scores \emph{when}
a system speaks rather than what it chooses: a question counts as
answered only if the response is produced within a time tolerance of
the annotated moment. It is the task that tests the response gate most
directly. Because it measures timing rather than answer accuracy like
the other seventeen tasks, we keep it out of the Contextual average.

Our reported value, 46.3 (95\% CI 39.1--53.7), comes from the run that
follows the official protocol, with a median timing error of 4.0\,s;
the gate stays silent on 15 of its 175 questions. Letting the model see
the full clip makes the gate fire on every question but leaves the
score unchanged.

Against the other systems --- Claude 3.5 Sonnet 64.7, GPT-4o 56.9,
Gemini 1.5 Pro 45.1, and every open model at 40.9 or below --- our
gate is indistinguishable from Gemini and clearly behind the two
strongest proprietary models. The confidence intervals are wide for
every method because the task has few questions per model, so we treat
Proactive Output as a diagnostic rather than as a headline result.


\clearpage
\section{Additional OVO-Bench Analysis}
\label{app:ovo_extra}
\label{app:ovo_breakdown}
\suppressfloats[t]

To keep the main comparison legible, we split OVO-Bench into its
Real-Time Visual Perception and temporal-reasoning task groups. OCR
denotes Optical Character Recognition; ACR, Action Recognition; ATR,
Attribute Recognition; STU, Spatial Understanding; FPD, Future
Prediction; and OJR, Object Recognition. The row named LLaVA-Video-7B
is the model listed as LLaVA-NeXT-Video-7B in the first release of the
benchmark. The StreamBridge row is its Qwen2-VL-7B + Stream-IT
configuration; its LLaVA-OV-7B backbone reaches 61.64 real-time and
48.13 overall here, and its StreamingBench results for both backbones
are given in Tables~\ref{tab:streamingbench_rtvu_detailed}
and~\ref{tab:streamingbench_other_detailed}. Human scores come from
watching the video at its native frame rate (24--30\,fps).

OVO-Bench has been revised since its first release, and the revision
changed most baseline scores: Gemini 1.5 Pro, for instance, moves from
70.77/62.32/57.15/65.25 in v1 to 69.32/62.54/57.15/63.00 in v2, and
Flash-VStream from 29.86 to 28.37 real-time average. Several method
papers still reproduce the v1 leaderboard. All reference rows in the
tables below are taken from the current v2 release of
\citet{niu2025ovobench} so that they are mutually consistent; rows
attributed to a method paper keep that paper's own reported values.
Note also that the TimeChat-Online row here is its 15.2\% keep-rate
configuration, whereas the StreamingBench tables use its 100\%
keep-rate configuration, because those are the configurations each
benchmark's comparison reports.

\subsection{Real-Time Visual Perception}
\suppressfloats[t]

\begin{table}[!htbp]
\caption{\textbf{OVO-Bench Real-Time Visual Perception, per task
(accuracy, \%).} The second block lists reference models rather than
streaming systems, from the current (v2) release of
\citet{niu2025ovobench}; Qwen3-VL-8B and MiniCPM-o~4.5 are from
\citet{lu2026aura}.}
\label{tab:ovo_realtime_detailed}
\centering
\scriptsize
\setlength{\tabcolsep}{2.6pt}
\renewcommand{\arraystretch}{1.10}
\begin{tabular}{lcccccccc}
\toprule
\textbf{Method} & \textbf{Frames} & \textbf{OCR} & \textbf{ACR} &
\textbf{ATR} & \textbf{STU} & \textbf{FPD} & \textbf{OJR} & \textbf{Avg.} \\
\midrule
Human & native & 94.0 & 92.6 & 94.8 & 92.7 & 91.1 & 94.0 & 93.20 \\
\midrule
Gemini 1.5 Pro~\citep{team2024gemini15} & 1 fps & 85.91 & 66.97 & 79.31 & 58.43 & 63.37 & 61.96 & 69.32 \\
GPT-4o~\citep{openai2024gpt4o} & 64 & 69.80 & 64.22 & 71.55 & 51.12 & 70.30 & 59.78 & 64.46 \\
LLaVA-Video-7B~\citep{zhang2024llavavideo} & 64 & 69.13 & 58.72 & 68.83 & 49.44 & 74.26 & 59.78 & 63.52 \\
LLaVA-OneVision-7B~\citep{li2024llavaonevision} & 64 & 66.44 & 57.80 & 73.28 & 53.37 & 71.29 & 61.96 & 64.02 \\
Qwen2-VL-7B~\citep{wang2024qwen2vl} & 64 & 60.40 & 50.46 & 56.03 & 47.19 & 66.34 & 55.43 & 55.98 \\
InternVL2-8B~\citep{chen2024internvl2} & 64 & 67.11 & 60.55 & 63.79 & 46.07 & 68.32 & 56.52 & 60.39 \\
LongVU-7B~\citep{shen2024longvu} & 1 fps & 53.69 & 53.21 & 62.93 & 47.75 & 68.32 & 59.78 & 57.61 \\
Qwen3-VL-8B-Instruct~\citep{bai2025qwen3vl} & 1 fps & 54.5 & 58.7 & 68.5 & 70.3 & 68.1 & 49.5 & 61.60 \\
\midrule
VideoLLM-Online~\citep{chen2024videollmonline} & 1 fps & 8.05 & 23.85 & 12.07 & 14.04 & 45.54 & 21.20 & 20.79 \\
Flash-VStream~\citep{zhang2024flashvstream} & 1 fps & 24.2 & 29.4 & 28.5 & 33.7 & 25.7 & 28.8 & 28.37 \\
Dispider~\citep{qian2025dispider} & 1 fps & 57.7 & 49.5 & 62.1 & 44.9 & 61.4 & 51.6 & 54.55 \\
TimeChat-Online~\citep{yao2025timechatonline} & 1 fps & 69.8 & 48.6 & 64.7 & 44.9 & 68.3 & 55.4 & 58.60 \\
StreamForest~\citep{zeng2025streamforest} & 1 fps & 68.46 & 53.21 & 71.55 & 47.75 & 65.35 & 60.87 & 61.20 \\
StreamBridge~\citep{wang2025streambridge}, Qwen2-VL-7B + Stream-IT & 1 fps & 84.56 & 71.56 & 74.14 & 49.44 & 75.25 & 72.83 & 71.30 \\
QueryStream~\citep{zhang2026querystream} & 1 fps & 74.5 & 47.7 & 70.7 & 46.6 & 71.3 & 57.6 & 61.40 \\
StreamAgent~\citep{yang2025streamagent} & 1 fps & 71.2 & 53.2 & 63.6 & 53.9 & 67.3 & 58.7 & 61.30 \\
MiniCPM-o 4.5-9B~\citep{cui2026minicpmo45realtimefullduplex} & 1 fps & 73.2 & 65.1 & 75.0 & 59.0 & 69.3 & 62.5 & 67.30 \\
\midrule
\textsc{Foresight} & adaptive & 85.91 & 71.27 & 75.88 & 64.97 & 78.81 & 73.25 & 75.02 \\
\bottomrule
\end{tabular}
\end{table}

\subsection{Backward Tracing and Forward Active Responding}
\suppressfloats[t]

EPM denotes Episodic Memory; ASI, Action Sequence Identification; HLD,
Hallucination Detection; REC, Repetition Event Count; SSR, Sequential
Steps Recognition; and CRR, Clues Reveal Responding.

\begin{table}[!htbp]
\caption{\textbf{OVO-Bench temporal-reasoning breakdown (accuracy,
\%).} Reference rows follow the same provenance as
Table~\ref{tab:ovo_realtime_detailed}, except VideoLLM-Online's Forward
Active Responding and overall, which the original release does not
report and which come from our own run.}
\label{tab:ovo_temporal_detailed}
\centering
\scriptsize
\setlength{\tabcolsep}{3.2pt}
\renewcommand{\arraystretch}{1.10}
\begin{tabular}{lccccccccc}
\toprule
& \multicolumn{4}{c}{\textbf{Backward Tracing}} &
\multicolumn{4}{c}{\textbf{Forward Active Responding}} & \\
\cmidrule(lr){2-5}\cmidrule(lr){6-9}
\textbf{Method} & \textbf{EPM} & \textbf{ASI} & \textbf{HLD} & \textbf{Avg.} &
\textbf{REC} & \textbf{SSR} & \textbf{CRR} & \textbf{Avg.} & \textbf{Overall} \\
\midrule
Human & 92.6 & 93.0 & 91.4 & 92.30 & 95.5 & 89.7 & 93.6 & 92.90 & 92.80 \\
\midrule
Gemini 1.5 Pro~\citep{team2024gemini15} & 58.59 & 76.35 & 52.64 & 62.54 & 35.53 & 74.24 & 61.67 & 57.15 & 63.00 \\
GPT-4o~\citep{openai2024gpt4o} & 57.91 & 75.68 & 48.66 & 60.75 & 27.58 & 73.21 & 59.40 & 53.40 & 59.54 \\
LLaVA-Video-7B~\citep{zhang2024llavavideo} & 56.23 & 57.43 & 7.53 & 40.40 & 34.10 & 69.95 & 60.42 & 54.82 & 52.91 \\
LLaVA-OneVision-7B~\citep{li2024llavaonevision} & 54.21 & 55.41 & 21.51 & 43.71 & 25.64 & 67.09 & 58.75 & 50.50 & 52.74 \\
Qwen2-VL-7B~\citep{wang2024qwen2vl} & 47.81 & 35.48 & 56.08 & 46.46 & 31.66 & 65.82 & 48.75 & 48.74 & 50.39 \\
InternVL2-8B~\citep{chen2024internvl2} & 48.15 & 57.43 & 24.73 & 43.44 & 26.50 & 59.14 & 54.14 & 46.60 & 50.15 \\
LongVU-7B~\citep{shen2024longvu} & 40.74 & 59.46 & 4.84 & 35.01 & 12.18 & 69.48 & 60.83 & 47.50 & 46.71 \\
Qwen3-VL-8B-Instruct~\citep{bai2025qwen3vl} & 58.8 & 41.1 & 23.1 & 41.00 & 21.2 & 51.3 & 40.4 & 37.70 & 46.80 \\
\midrule
VideoLLM-Online~\citep{chen2024videollmonline} & 22.22 & 18.80 & 12.18 & 17.73 & 11.20 & 44.47 & 38.80 & 29.00 & 21.12 \\
Flash-VStream~\citep{zhang2024flashvstream} & 39.1 & 37.2 & 5.9 & 27.38 & 8.0 & 67.3 & 60.0 & 45.09 & 33.61 \\
Dispider~\citep{qian2025dispider} & 48.5 & 55.4 & 4.3 & 36.06 & 18.1 & 37.4 & 48.8 & 34.72 & 41.78 \\
TimeChat-Online~\citep{yao2025timechatonline} & 53.9 & 62.8 & 9.1 & 42.00 & 32.5 & 36.5 & 40.0 & 36.40 & 45.60 \\
StreamForest~\citep{zeng2025streamforest} & 58.92 & 64.86 & 32.26 & 52.02 & 32.81 & 70.59 & 57.08 & 53.49 & 55.57 \\
StreamBridge~\citep{wang2025streambridge}, Qwen2-VL-7B + Stream-IT & 67.68 & 57.43 & 79.03 & 68.05 & 19.17 & 64.25 & 61.67 & 48.36 & 62.57 \\
QueryStream~\citep{zhang2026querystream} & 54.2 & 63.5 & 8.6 & 42.10 & 33.2 & 43.1 & 40.8 & 39.03 & 47.51 \\
StreamAgent~\citep{yang2025streamagent} & 54.8 & 58.1 & 25.8 & 41.70 & 35.9 & 48.5 & 52.0 & 45.40 & 49.40 \\
MiniCPM-o 4.5-9B~\citep{cui2026minicpmo45realtimefullduplex} & 55.2 & 66.9 & 45.7 & 55.90 & 43.0 & 66.0 & 58.8 & 55.90 & 59.70 \\
\midrule
\textsc{Foresight} & 62.2 & 69.5 & 33.6 & 55.10 & 32.1 & 77.2 & 59.8 & 56.37 & 62.16 \\
\bottomrule
\end{tabular}
\end{table}


\clearpage
\section{Anticipation Analysis}
\label{app:anticipation_lead_time}
\suppressfloats[t]

\begin{figure}[!htbp]
    \centering
    \includegraphics[width=0.9\linewidth]{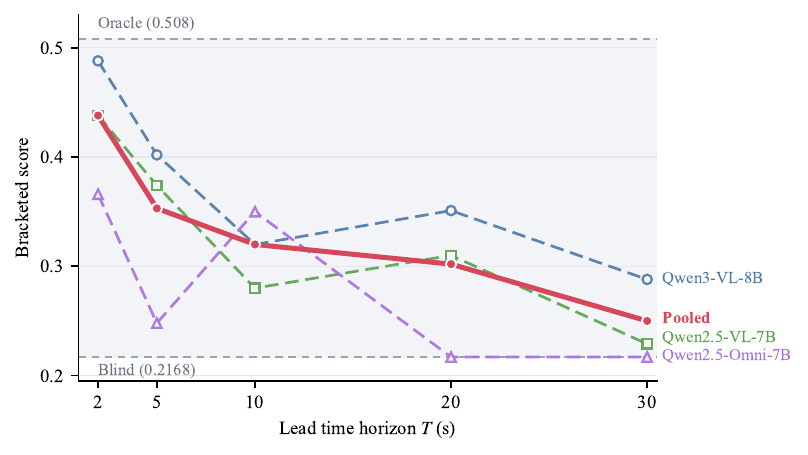}
    \caption{\textbf{Anticipation quality against lead time.} Each curve
    shows how well a frozen VLM anticipates an upcoming event when it is
    asked $T$ seconds before the event, using only the stream seen so far.
    Scores are bracketed between two references: \emph{Blind} (0.2168), a
    baseline that never sees the video, and \emph{Oracle} (0.508), which
    observes the event itself. Dashed lines are individual backbones and the
    solid red line pools them. The pooled score is well above the blind
    baseline for near-future events and falls steadily as the horizon
    grows, approaching it by $T=30$\,s; individual backbones are noisier,
    and Qwen2.5-Omni-7B reaches the blind baseline from $T=20$\,s. Anticipation
    is therefore most reliable over the short horizons at which
    \textsc{Foresight} schedules its next check.}
    \label{fig:llm_anticipation}
\end{figure}

This appendix provides the direct empirical evidence for the anticipatory
capability used by \textsc{Foresight} in Section~\ref{sec:method}.
Figure~\ref{fig:llm_anticipation} measures how that capability changes with
the amount of time remaining before an event.

This experiment is what led us to \textsc{Foresight}. We asked a simple
question: if a frozen VLM watches a stream and is stopped $T$ seconds
before an event, can it already tell that the event is coming? If it
can, the model does not need to be asked about every frame. It can
decide for itself when the next moment worth looking at will arrive,
and spend its computation there. Figure~\ref{fig:llm_anticipation}
measures this: at each lead time $T$, the model sees the stream up to
$T$ seconds before an event and must anticipate it from that context
alone.

\paragraph{Anticipation is real, but short-lived.}
The answer is yes, for the near future. Two seconds before an event,
the pooled score closes about three quarters of the gap between the
blind baseline and the oracle, so the model already sees much of what
is about to happen. By 30\,s that share has fallen to about a tenth.
This matches intuition: a player dribbling into the box makes a shot in
the next few seconds likely, but says little about the score half a
minute later. What a model can infer about the future comes from cues
that are visible now, and those cues stop being informative as the
horizon grows.

\paragraph{Why the controller must check often.}
This decay shapes how \textsc{Foresight} plans. Each plan is a short
bet on the immediate future: it says when to look again, what to look
for, and how densely to sample until then. Because the model's view of
the future is reliable only a few seconds ahead, those bets must be
renewed often. A controller that checked once and scheduled the next
check far ahead would be acting on a prediction it can no longer
support, and would miss events that its last look could not foresee.
\textsc{Foresight} therefore keeps checking at a steady, scene-dependent
rate, lengthening $\Delta_t$ when nothing is developing and shortening
it when something is. Keeping those checks cheap is the subject of
Appendix~\ref{app:efficiency}.

\paragraph{Fewer checks is an open problem.}
The same result marks the limit of our approach. The ideal controller
would check about once per event, and a system that could plan
reliably tens of seconds ahead would come close. Current VLMs cannot:
their foresight fades within seconds, so a controller built on them
must spend some checks on moments where nothing happens. Reaching the
ideal rate of one check per event, or even a few checks per minute on
calm streams, would need a model whose anticipation lasts much longer
than today's.

\paragraph{Training for anticipation.}
We see this as a property of how VLMs are trained rather than a fixed
limit. Pretraining teaches a model to describe what is in front of it,
not to predict what comes next, so any foresight it has is a by-product.
An objective that asks the model to predict upcoming events, their
timing, or the next useful moment to look would make anticipation part
of what the model learns directly. Such a model would be proactive by
nature, and a controller like \textsc{Foresight} could then plan further
ahead with fewer checks.


\clearpage
\section{Inference Efficiency}
\label{app:efficiency}
\suppressfloats[t]

\textsc{Foresight} is designed so that thinking about the stream costs
little beyond watching it. We make this precise with a simple cost
model and show where each design choice enters it.

\paragraph{Setup.}
Consider a stream of length $L$ seconds in which $N$ frames
$x_1,\dots,x_N$ are admitted, each encoded into $m$ visual tokens, and
which contains $E$ events of interest. The Think LLM runs at check times
$t_1<\dots<t_K$ with $t_{k+1}=t_k+\Delta_{t_k}$, so its average check
rate is $q=K/L$. We measure work $\mathcal{W}$ in tokens processed by
the VLM; with $P$ parameters, a dense forward pass costs about $2P$
FLOPs per token, so compute is $2P\,\mathcal{W}$.

\paragraph{The baseline: re-reading the stream.}
A controller without a shared cache must rebuild its context at every
check. At check $k$ it prefills the whole state $C_{t_k}$, which holds
about $m\,n_k$ tokens for the $n_k$ frames seen so far, plus the prompt
of length $|p|$, before writing a plan of $|\pi|$ tokens:
\begin{equation}
    \mathcal{W}_{\text{re-read}}
    \;=\; \sum_{k=1}^{K}\bigl(m\,n_k + |p| + |\pi|\bigr)
    \;=\; \mathcal{O}\!\left(K\,m\,N\right).
    \label{eq:cost_reread}
\end{equation}
Every check pays again for the whole history. With evenly spaced checks
$K\propto L$ and $N\propto L$, so this cost grows as $L^2$: doubling the
stream quadruples the work.

\paragraph{\textsc{Foresight}.}
\textsc{Foresight} splits the same work into what must be paid once and
what is paid per decision:
\begin{equation}
    \mathcal{W}_{\textsc{Foresight}}
    \;=\; \underbrace{\rho\, m\, N_{\!\text{adapt}}}_{\text{ingest, once}}
    \;+\; \underbrace{K\,c_{\text{probe}}}_{\text{checks}}
    \;+\; \underbrace{\textstyle\sum_{j} |u_{t_j}| + |s_{t_j}|}_{\text{plans at evidence}} ,
    \label{eq:cost_foresight}
\end{equation}
where the last sum runs over the steps $j$ at which the probe finds
enough evidence to write a plan. Each term corresponds to one design
choice.

\begin{itemize}
    \item \emph{Shared KV cache.} Each visual token is prefilled exactly
    once, when the Ingest LLM appends it to $C_t$. The Think LLM reads a
    snapshot $C_{t_k}^{\mathrm{snap}}$ of that cache instead of
    rebuilding it, so the $m\,n_k$ term of Eq.~\ref{eq:cost_reread}
    disappears from every check, and the prompt $p$, prefilled once as a
    persistent prefix, disappears with it.
    \item \emph{Adaptive frames and pruning.} The plan admits
    $N_{\!\text{adapt}}\le N$ frames, sampling densely only around
    expected events, and Keep and Ignore Classes retain a fraction
    $\rho\le 1$ of each frame's tokens. Both shrink the one-time ingest
    term, which is the largest in Eq.~\ref{eq:cost_foresight}.
    \item \emph{The probe inside the plan.} Most checks find nothing
    new. For those, the Think LLM decodes only the first field of the
    plan and reads \texttt{have\_enough\_info} from a single step of
    logits, so a check costs a constant $c_{\text{probe}}$ of a few
    tokens rather than a full plan.
    \item \emph{Diff-based plans.} When evidence does arrive, the model
    writes only the fields that change, $|u_{t_j}|+|s_{t_j}|\ll|\pi|$,
    and the merge $U\leftarrow\operatorname{merge}(U,u_t)$ keeps the
    rest.
\end{itemize}

\paragraph{What this buys.}
Two properties follow. First, the controller's cost grows with the
number of decisions, not with the length of the video: the check term
is linear in $K$ with a small constant, and the plan term is roughly
proportional to the number of events $E$. Second, the cost of each
decision does not grow over time. A decoded token attends to the cache,
so its cost scales with $|C_t|$; compaction and eviction keep $|C_t|$
within a fixed budget $B$, so planning at the end of a long stream is as
cheap as at the start. The one-time ingest is the cost of seeing the
video at all, and no controller can avoid it.

A natural lower bound is the \emph{oracle}, a controller with the same
ingest that knows in advance when the $E$ events occur and plans exactly
once at each of them:
\begin{equation}
    \mathcal{W}_{\text{oracle}}
    \;=\; \rho\, m\, N_{\!\text{adapt}} \;+\; \sum_{j=1}^{E} |u_{t_j}|+|s_{t_j}| .
    \label{eq:cost_oracle}
\end{equation}
Subtracting Eq.~\ref{eq:cost_oracle} from Eq.~\ref{eq:cost_foresight}
leaves only the price of not knowing the future,
\begin{equation}
    \mathcal{W}_{\textsc{Foresight}} - \mathcal{W}_{\text{oracle}}
    \;\approx\; K\,c_{\text{probe}} ,
    \label{eq:cost_excess}
\end{equation}
a few tokens per check. \textsc{Foresight} keeps this term small in two
ways: $c_{\text{probe}}$ is tiny, and $K$ adapts to the video, because
the plan lengthens $\Delta_t$ when nothing relevant is happening.

\paragraph{Why the design choices matter.}
Each saving above is a design choice, and giving up any one of them is
costly. Under the cost model, at one check every four seconds, dropping
the shared KV cache raises total compute about $40\times$, since every
check re-reads the stream; sampling at a fixed rate without pruning
roughly triples it by inflating the ingest term. The other choices
govern the controller's overhead above the oracle: a full plan instead
of a diff makes it about $25\times$ larger, a diff at every check
without a probe about $4\times$, and a probe run as a separate pass
about $2\times$. These factors come from the model, not from
measurement, but their ordering follows from the structure of
Eq.~\ref{eq:cost_foresight}.

\begin{figure}[!htbp]
    \centering
    \includegraphics[width=0.8\linewidth]{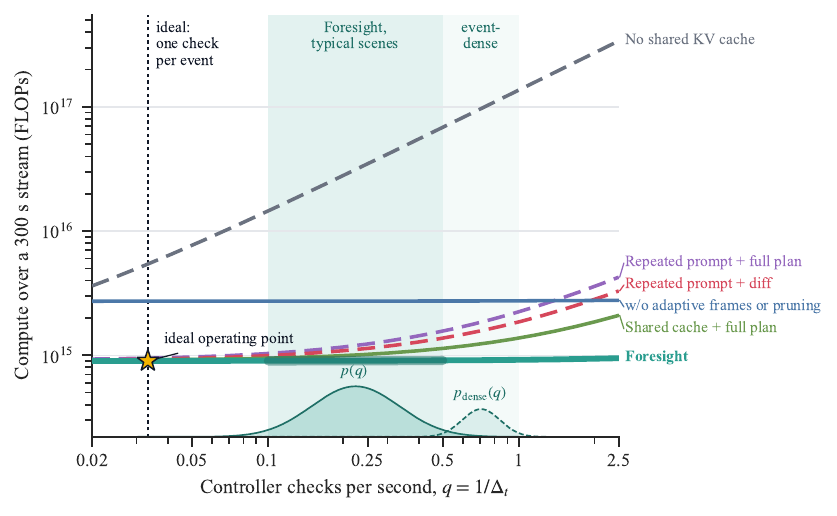}
    \caption{\textbf{Total compute against check rate.} Each curve is the
    compute a design spends on a 300\,s stream. Designs that re-read the
    stream at every check grow quickly with $q$: without a shared KV
    cache every check prefills the whole history again, and with a
    repeated prompt every check pays for the prompt on top of the plan.
    Ingesting at a fixed rate without pruning costs the same at every $q$,
    so it appears as a flat line set by how many frames are admitted.
    \textsc{Foresight} stays almost flat across its operating region:
    nearly all of its cost is the one-time ingest of the video, and each
    check adds only a short probe. The shaded bands and the distributions
    $p(q)$ and $p_{\mathrm{dense}}(q)$ show where its check rate falls in
    typical and event-dense scenes; the star marks the ideal operating
    point of one check per event.}
    \label{fig:operating_point_1}
\end{figure}

\begin{figure}[!htbp]
    \centering
    \includegraphics[width=0.8\linewidth]{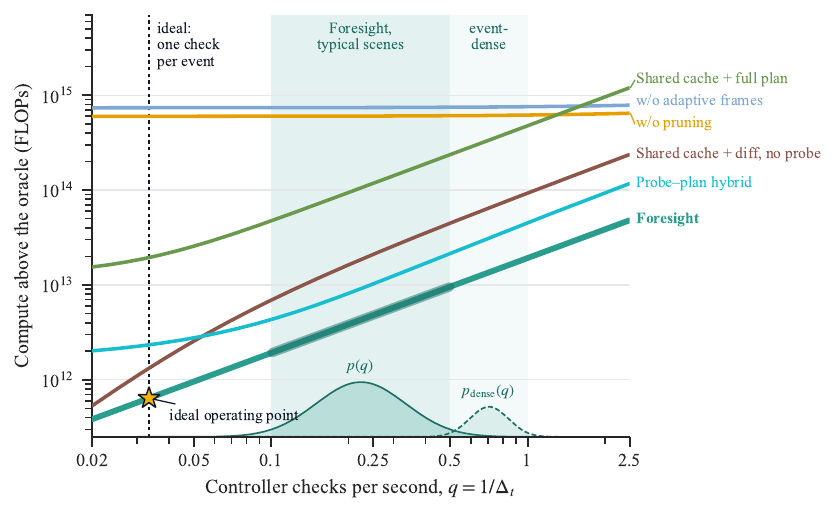}
    \caption{\textbf{Compute above the oracle.} The same model, with the
    oracle's compute subtracted from every design. Because the oracle
    shares the ingest pipeline, what remains is only the work a design
    spends beyond the least it could, which makes differences that are
    invisible in Figure~\ref{fig:operating_point_1} easy to read. Removing
    adaptive frames or pruning adds a fixed cost from extra visual
    tokens, so these curves are flat. The controller variants grow with
    $q$, and their order shows what each saving buys: a full plan at
    every check costs most, then a diff at every check, then a separate
    probe followed by a plan, and least of all \textsc{Foresight}, which
    folds the probe into the plan and writes a diff only when evidence
    arrives. In its operating region \textsc{Foresight} wastes about half
    as much as the probe--plan hybrid, a quarter as much as a diff at
    every check, and about a hundredth as much as the designs without
    adaptive frames or pruning.}
    \label{fig:operating_point_2}
\end{figure}

\paragraph{Where \textsc{Foresight} operates.}
Figures~\ref{fig:operating_point_1} and~\ref{fig:operating_point_2} plot
the cost model against the check rate $q=1/\Delta_t$, the number of
planning steps per second of video. The oracle of
Eq.~\ref{eq:cost_oracle} checks once per event, so its rate $q=E/L$ is
the \emph{ideal operating point}. No real controller can sit there,
because knowing when an event will happen is exactly what a streaming
model has to work out. \textsc{Foresight} instead chooses $\Delta_t$
from what it sees, so its check rate is a property of the video rather
than a fixed setting: calm, slowly changing scenes let it wait several
seconds, while fast motion, many objects or a quickly developing goal
pull the next check closer. We therefore draw its operating point as a
distribution: most streams fall in a window of one check every 2 to
10\,s, and event-dense streams reach one check per second.

Together the figures make one point from two sides: in total compute
\textsc{Foresight} is barely distinguishable from the oracle, and in the
extra compute it does spend it is the cheapest design we consider. The
curves use representative token counts, and the distributions $p(q)$
and $p_{\mathrm{dense}}(q)$ are assumed; measured check rates on the
benchmarks would replace them.

\subsection{Measured Cost of the Streaming Baselines}
\label{app:baseline_cost}
\suppressfloats[t]

Online task scores say nothing about whether a system can keep pace
with the stream it is watching. The natural reference is the video
itself: a real-time system must consume one second of footage in at
most one second of compute. We therefore record, for the baselines we
ran ourselves, the real-time (RT) factor, defined as wall-clock
seconds of compute per second of video. All measurements use a single
GH200 per sample on the same OmniPro Online set (2,700 videos,
143 hours of footage), and the wall clock includes the shared
per-frame decoding and vision-tower cost attributed back to each
sample.

Neither StreamAgent arm runs in real time. With every mechanism of the
method active, each second of video costs roughly two to three seconds
of GH200 time --- 2.13$\times$ with the omni backbone and 2.70$\times$
with the vision-only one --- so on this hardware it falls behind the
stream. The gap is structural, since
every clip triggers a caption, two candidate plans, a plan-scoring
pass and a trigger check before any answer is produced. The probe
configuration, which drops per-clip planning, plan scoring and the
trigger gate and keeps only captioning plus two answerer calls per
ground-truth trigger, runs at 0.58$\times$ -- faster than the stream,
at 26\% of the online cost.

QueryStream sits at the opposite end of the same axis. Measured over
every scored OmniPro sample, it averages 0.053$\times$ on its
Qwen2.5-VL-7B backbone and 0.084$\times$ on Qwen3-VL-8B, so both
finish well inside real time, and the newer backbone is consistently
the more expensive of the two. As for StreamAgent, the ratio is close
to flat across video-length buckets, confirming that cost scales with
duration rather than with anything specific to long videos.


\clearpage
\section{Controller Configuration and Code Release}
\label{app:controller_config}
\suppressfloats[t]

Table~\ref{tab:controller_config} lists the response-gate settings used
on OmniPro Online, fitted per task on the 20\% calibration split and
held fixed on the 80\% evaluation split.

\begin{table}[H]
\caption{\textbf{Response-gate settings on OmniPro Online.} Threshold
on $p_{\mathrm{hit},t}$, gate mode, and refractory interval for each
task, fitted on the calibration split.}
\label{tab:controller_config}
\centering
\small
\renewcommand{\arraystretch}{1.10}
\begin{tabular}{llccc}
\toprule
\textbf{Task} & \textbf{Abbr.} & \textbf{Threshold} & \textbf{Mode} &
\textbf{Refractory (s)} \\
\midrule
Target-Ground      & TG & 0.50  & edge  & 300 \\
Event-Alert        & EA & 0.45  & edge  & 600 \\
State-Monitor      & SM & 0.80  & level & 7   \\
Conditional-Alert  & CA & 0.98  & level & 10  \\
Step-Instruction   & SI & 0.01  & level & 7   \\
Snapshot-Count     & SC & 0.985 & edge  & 600 \\
\bottomrule
\end{tabular}
\end{table}

Each setting has three parts.
The \emph{threshold} is the evidence confidence $p_{\mathrm{hit},t}$ at
which the gate may fire. The \emph{mode} decides how often it fires: an
\emph{edge} gate fires once when the confidence first crosses the
threshold, and a \emph{level} gate fires whenever the confidence stays
above it. The \emph{refractory interval} is the shortest time, in
seconds, allowed between two responses, which stops a lasting event
from being reported again. These values come from the video-only
configuration used for all reported results. The settings follow the
nature of each task: alerts and counts that should be reported once per
event use an edge gate with a long refractory interval, while tasks that
track a changing state, such as State-Monitor and Step-Instruction, use
a level gate with a short one so that every new step or state change
can be reported.

\paragraph{Code and materials.}
Code, prompts, and evaluation scripts are provided as anonymized
supplementary material and will be released publicly upon publication.

\end{document}